\documentclass[authorversion, acmsmall]{acmart}

\let\digamma\relax

\AtBeginDocument{%
  }

\setcopyright{cc}
\setcctype{by}
\acmJournal{PACMCGIT}
\acmYear{2025} \acmVolume{8} \acmNumber{1} \acmArticle{12} \acmMonth{5} \acmDOI{10.1145/3728293}

\usepackage[utf8]{inputenc}
\usepackage{amsmath, amssymb, amsfonts, mathtools}
\usepackage{bm, pifont}

\usepackage{graphicx}
\usepackage{svg}
\usepackage{subcaption}
\usepackage{tikz}
\usepackage{pgf}
\usepackage{pgfplots}
\usepackage{tikzscale}
\usetikzlibrary{arrows, arrows.meta, external, calc, positioning, spy}

\usepackage{multirow}
\usepackage{tabularx}
\usepackage{colortbl}

\usepackage{xcolor}
\usepackage{color}
\usepackage{soul}
\usepackage{nicefrac}

\usepackage{url}
\usepackage[capitalize]{cleveref}
\crefname{section}{Sec.}{Secs.}
\Crefname{section}{Section}{Sections}
\Crefname{table}{Table}{Tables}
\crefname{table}{Tab.}{Tabs.}
\usepackage{xspace} 

\Crefformat{figure}{#2Fig.~#1#3}
\Crefmultiformat{figure}{Figs.~#2#1#3}{ and~#2#1#3}{, #2#1#3}{ and~#2#1#3}

\definecolor{colorTrd}{rgb}{0.95,0.95,0.65}
\definecolor{colorSnd}{rgb}{1, 0.85, 0.7}
\definecolor{colorFst}{rgb}{1, 0.7, 0.7}
\definecolor{green}{rgb}{0.45, 0.62, 0.31}
\definecolor{blue}{rgb}{0.42, 0.60, 0.82}
\definecolor{yellow}{rgb}{0.80, 0.64, 0.22}

\newcommand{\noi}{\noindent}
\newcommand{\point}{\textbf{x}}
\newcommand{\real}{\mathbb{R}}

\newcommand{\rthree}{\real^3}
\newcommand{\image}{I}
\newcommand{\images}{\mathcal{\image}}
\newcommand{\nimages}{n}
\newcommand{\sdf}{f}
\newcommand{\npoints}{{n_p}}
\newcommand{\depthmap}{D}

\newcommand{\ie}{\emph{i.e., }}
\newcommand{\eg}{\emph{e.g., }}
\newcommand{\etal}{\emph{et al.}}
\newcommand{\viewdir}{v}
\newcommand{\ray}{r}
\newcommand{\stwo}{\mathcal{S}^2}
\newcommand{\networkparams}{\Theta}
\newcommand{\normalvec}{\mathbf{n}}

\usepackage{acmart-taps}

\begin{document}

\title{Normal-guided Detail-Preserving Neural Implicit Function for High-Fidelity 3D Surface Reconstruction}

\author{Aarya Patel}
\email{aaryap@iiitd.ac.in}
\affiliation{%
  \institution{Graphics Research Group, IIIT Delhi}
  \city{New Delhi}
  \state{Delhi}
  \country{India}
}

\author{Hamid Laga}
\email{h.laga@murdoch.edu.au}
\affiliation{%
  \institution{Murdoch University}
  \city{Perth}
  \state{Western Australia}
  \country{Australia}
}

\author{Ojaswa Sharma}
\email{ojaswa@iiitd.ac.in}
\affiliation{%
  \institution{Graphics Research Group, IIIT Delhi}
  \city{New Delhi}
  \state{Delhi}
  \country{India}
}

\renewcommand{\shortauthors}{Aarya Patel, Hamid Laga, and Ojaswa Sharma}

\begin{abstract}
Neural implicit representations have emerged as a powerful paradigm for 3D reconstruction. However, despite their success, existing methods fail to capture fine geometric details and thin structures, especially in scenarios where only sparse multi-view RGB images of the objects of interest are available. This paper shows that training neural representations with first-order differential properties (surface normals) leads to highly accurate 3D surface reconstruction, even with as few as two RGB images. Using input RGB images, we compute approximate ground-truth surface normals from depth maps produced by an off-the-shelf monocular depth estimator. During training, we directly locate the surface point of the SDF network and supervise its normal with the one estimated from the depth map. Extensive experiments demonstrate that our method achieves state-of-the-art reconstruction accuracy with a minimal number of views, capturing intricate geometric details and thin structures that were previously challenging to capture. The source code and additional results are available at \href{https://graphics-research-group.github.io/sn-nir/}{https://graphics-research-group.github.io/sn-nir}.

\end{abstract}

\ccsdesc[500]{Computing methodologies~Reconstruction}
\keywords{Neural Rendering, Implicit Modeling, 3D Shape Reconstruction}


\maketitle
\section{Introduction}
\label{sec:intro}

Accurate 3D surface reconstruction from partial observations like RGB images is a challenging problem in computer vision and graphics that has been extensively investigated. Early learning-based methods relied on discrete surface representations in the form of polygonal meshes~\cite{poisson,Labatut_2007,Lorensen1987Marching}, point clouds~\cite{attene2000automatic, Kolluri, huang2009consolidation, livny2010automatic, nan2017polyfit}, or voxel grids~\cite{curless1996volumetric, seitz1999photorealistic}, which often struggle to capture complex topologies and fine-grained geometric details while maintaining memory efficiency. In recent years, neural implicit representations have emerged as a promising alternative, representing surfaces as continuous decision boundaries of neural networks while offering several advantages such as memory efficiency, resolution-agnostic output, and the ability to represent complex topologies.

Despite their impressive capabilities and performance, existing neural implicit reconstruction approaches \cite{park2019deepsdf, chibane2020neural, wang2021neus, yariv2021volume} face several limitations. They often require training on a large number of views and struggle to capture fine geometric details, particularly in scenarios with sparse multi-view inputs or complex scene geometries. This challenge arises from the difficulty in accurately reconstructing intricate surface features from limited and potentially noisy observations. The self-supervised nature of these methods, where they learn to map 3D coordinates to geometric representations without explicit ground truth, contributes to this difficulty. At inference time, while the learned neural implicit function can be queried at arbitrary 3D locations to reconstruct the complete surface geometry, the quality of fine details and thin structures may be compromised, especially when trained on limited input views.

This work explores the benefits of incorporating high-order differential properties, such as surface normals, into neural implicit surface reconstruction to address the limitations of existing approaches. Current methods often require a large number of input views and struggle with fine geometric details, especially in scenarios with sparse multi-view inputs or complex geometries. Our key insight is that high-order differential properties carry richer information about local surface geometry than zero-order properties such as surface points. By explicitly integrating these properties into the neural network's supervision, we can significantly improve the network's ability to capture intricate surface details and generate high-quality reconstructions, even with limited input views.

Our method differs from current normal integration methods such as NeuRIS~\cite{wang2022neuris} and MonoSDF~\cite{Yu2022MonoSDF} in the way the surface point of the Signed Distance Field (SDF) network is calculated. Instead of using volumetric ray accumulation that brings inherent geometric errors in the estimated surface, we directly locate the zero level set of the SDF using our proposed approach, which is elaborated in the subsequent sections.
To train the network, we leverage an off-the-shelf monocular depth estimation network \cite{depthanything} to obtain the necessary depth information. These depth maps are then used to estimate the corresponding surface normal maps, providing the required geometric cues for our normal-guided reconstruction process. The main contributions of this paper can be summarized as follows:
\begin{itemize}
    \item We propose a novel neural implicit surface reconstruction method that is explicitly supervised using high-order differential properties, such as surface normals, directly estimated from the input RGB images.  

    \item We propose to estimate surface normals for supervision by directly locating the corresponding surface points on the zero level set of the SDF. This significantly helps achieve highly accurate surface reconstructions compared to the approach used in~\cite{Yu2022MonoSDF,wang2022neuris}.
    
    \item We conduct comprehensive qualitative and quantitative evaluations on challenging multi-view datasets, such as DiLiGent-MV~\cite{diligentMV}, BlendedMVS \cite{yao2020blendedmvs} and DTU \cite{jensen2014large} benchmarks. Our experimental results demonstrate that the proposed method significantly outperforms state-of-the-art techniques, achieving high-accuracy surface reconstructions, particularly in sparse multi-view stereo scenarios.

\end{itemize}

\section{Related work}
\label{sec:related_work}
\subsection{Neural surfaces from dense and sparse MVS}
\label{sec:related_work_mvs}

Early deep learning-based Multi-View Stereo (MVS)  methods~\cite{laga2022survey}  replace one or multiple blocks of the traditional MVS pipeline~\cite{furukawa2010,mvs_goesele} with deep neural networks. 
However, they still face challenges in recovering fine geometric details and reconstructing thin structures. 
Neural implicit functions~\cite{park2019deepsdf,mildenhall2021nerf} have emerged as a powerful, continuous, memory-efficient, and detail-preserving representation for 3D reconstruction from dense multiview images. 
The idea is to represent the surface of a 3D object as the zero level set of its SDF~\cite{park2019deepsdf}  and then train a Multi-Layer Perceptron (MLP) to learn a function that maps any point $\point\in \rthree$  to its SDF.
This powerful representation enables both 3D reconstruction and novel view synthesis~\cite{oechsle2021unisurf, yariv2020multiview, yariv2021volume, wang2021neus}. It has been extended in many ways to enable the representation and reconstruction of dynamic scenes~\cite{pumarola2021d} and large, unbounded scenes~\cite{gu2024ue4}, and improve the computation time \cite{oechsle2021unisurf,yariv2020multiview,yariv2021volume,mueller2022instant,wang2023neus2} and reconstruction accuracy~\cite{wang2021neus,darmon2022improving,li2023neuralangelo}. 


To enable 3D reconstruction from sparse RGB views, \eg as few as three RGB images per scene, SparseNeuS~\cite{long2022sparseneus} and VolRecon~\cite{ren2023volrecon} learn generalized geometric priors from images of a large number of scenes, and then fine-tune on new scenes.
Instead of using costly large training priors, NeuSurf~\cite{huang2024neusurf} exploits the prior of surface points, obtained using Structure-from-Motion (SfM) methods such as COLMAP~\cite{schonberger2016structure}.

Unlike previous works, our method does not rely on expensive large training priors, as in SparseNeuS~\cite{long2022sparseneus} and VolRecon~\cite{ren2023volrecon} and is a single-stage approach, unlike NeuSurf \cite{huang2024neusurf}. It attempts to exploit high-order differential properties of surfaces to learn a neural surface representation, achieving unprecedented results even when as few as two RGB images, with minimum overlap, are available. 
\subsection {Neural surfaces from multi-view photometric stereo}
\label{sec:neural_fields_mvps}
Traditional Multi-View Photometric Stereo (MVPS) methods \cite{lim2005passive,hernandez2008multiview, wu2010fusing, park2016robust, logothetis2019differential,li2020multi} first use MVS to reconstruct an approximate 3D shape and then employ Photometric Stereo (PS) to estimate the surface normals for each viewing direction, using images captured under varying lighting conditions.
Finally, they refine and improve the initial coarse 3D geometry by incorporating the surface normal information obtained through the PS step.  

Kaya \etal~\cite{kaya2022neural} was among the first to tackle the MVPS problem using NeRFs. However, it does not produce high-quality 3D geometry.
Kaya \etal~\cite{kaya2022uncertainty} improved~\cite{kaya2022neural} by taking advantage of an uncertainty-aware deep neural network that integrates MVS depth maps with PS normal maps to reconstruct 3D geometry.
PS-NeRF \cite{yang2022ps} introduced a neural inverse rendering method for MVPS under uncalibrated lights. The technique uses PS normal maps to regularize the gradient of UNISURF~\cite{oechsle2021unisurf} while relying on MLPs to explicitly model surface normals, BRDF, illumination and visibility. 
A recent work, RNb-NeuS~\cite{Brument23}, integrates multi-view reflectance and normal maps acquired through photometric stereo into the NeuS~\cite{wang2021neus} framework.
Another recent work, SuperNormal \cite{supernormal2024cao}, mainly focuses on the acceleration of training of MVPS methods by using the multi-resolution hash encoding technique to obtain good 3D reconstruction results.
However, current methods still require a large number of images with varying lighting per view. Also, as shown in \Cref{sec:results}, the reconstruction accuracy decreases significantly when using less than five views. Our proposed approach addresses these fundamental limitations by proposing a sparse multi-view stereo-based approach that only requires as few as two views.
\subsection{Neural surface reconstruction with normals}
Surface normals have previously been used as auxiliary information for neural surface reconstruction in NueRIS \cite{wang2022neuris} and MonoSDF \cite{Yu2022MonoSDF}. Both works approximate rendered surface normals observed from a particular viewpoint along a ray using volume accumulation given by 
\begin{equation}\label{eq:1}
    \mathbf{\hat{n}} = \sum_{i=1}^{\npoints} T_i\alpha_i \mathbf{n}_i,
\end{equation}

\noi where, $\mathbf{n}_i = \nabla_{\point_i} f(\point_i)$ is the gradient of the SDF $f$ evaluated at the point $\point_i$, $T_i$ denotes the transmittance accumulated up to point $\point_i$ and $\alpha_i$ denotes the alpha value of the sampled point. $\npoints$ is the number of points sampled along the ray. However, it is observed that applying standard volume rendering to accumulate the normals associated with the SDF leads to inherent geometric errors and a discernible geometric bias in the estimated surface~\cite{wang2021neus}.
Unlike~\cite{wang2022neuris, Yu2022MonoSDF}, in this paper, we directly locate the zero level set of the SDF network using our proposed surface extraction method that ensures that we focus on the true surface optimization.

\section{Method}
\label{sec:method}

\begin{figure}[t]
    \centering
    \includegraphics[width=\textwidth]{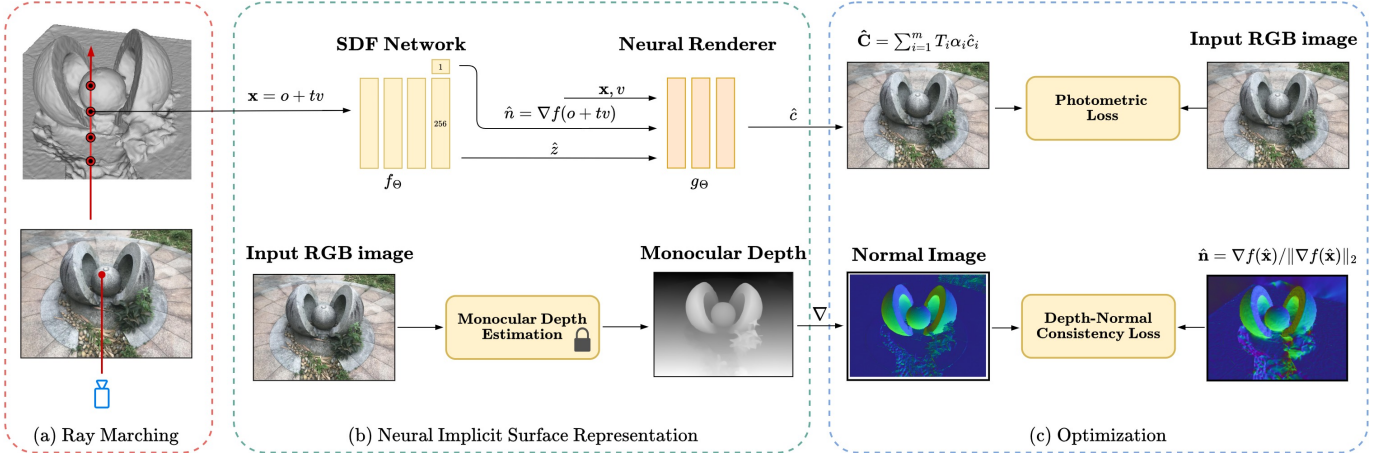}
    \caption{An overview of our method. Our method involves the following steps: (a) We cast camera rays from each pixel into the scene while sampling $m$ points along the ray. (b) We find the corresponding SDF value and rendered RGB color $\hat{c}$ for each sampled point using an SDF and Neural Renderer network, respectively. Then, we use volume rendering in a differentiable manner to find the rendered color for each pixel. (c) Finally, we additionally supervise our network with Depth-Normal Consistency loss, enabling 3D reconstruction of intricate geometric details and
thin structures.}
    \label{fig:method}
\end{figure}

\label{sec:oerview}

Given $\nimages$ calibrated multiview RGB images $\images = \{\image_i\}_{i=1}^{\nimages}$ of an object, we aim to train a function composed of two neural networks: the first known as the SDF Network $f_\theta$ that learns to map any point $\point \in \rthree$ to its signed distance field (SDF) and the second network, neural renderer $g_\theta$ that predicts the color value $\hat{c}$ as viewed from viewing direction $\viewdir \in \stwo$ at a point $\point$. The surface is then defined as the zero-level set of the SDF. 
Traditionally, this is optimized using a reprojection loss that involves an appearance (color) loss along with 2D Chamfer loss and regularization loss. However, these loss terms do not characterize the local geometry and local interactions between neighboring surface points. 
Thus, we introduce a depth-normal consistency loss term to explicitly enforce the normals of the estimated geometry to match the normals of the target 3D surface obtained from monocular depth images. This results in highly accurate 3D reconstructions.\Cref{fig:method} provides an overview of the proposed framework.


\subsection{Normal-guided neural surfaces}
\label{sec:method_normal_guided_NeuS} 
In this section, we introduce our neural surface representation of 3D objects, which is guided by the normal maps of their underlying surfaces. 
Vanilla Neural Radiance Fields (NeRF) \cite{mildenhall2021nerf} based methods use a set of images and corresponding camera poses to train an MLP network that takes a  3D point $\point \in\rthree$ and a 2D viewing direction $\viewdir \in \stwo$, and outputs the volume density $\sigma$  at $\point$ and its radiance as viewed from $\viewdir$. To render a pixel, a ray $\ray$ is cast from the camera center $o$ through the pixel. A 3D point along the ray $\ray$ through the corresponding pixel is represented as:
  \begin{equation}\label{eq:ray_eq}
     \point(t) = o + t\viewdir,  \text{ with }  t \geq 0.
 \end{equation}

\noi Here,  $t$ is the distance between $o$ and $\point$. NeRF samples $\npoints$ points $\{ \point_i \}_{i=1}^\npoints$ along the ray $\ray$. The density $\sigma_i$ and appearance $c_i$ are predicted by querying the network for the point $\point_i$ and direction $\viewdir= (\theta, \phi)$. Then, $\alpha$ compositing \cite{max1995optical} accumulates these values for each point along the ray $\ray$ to estimate the pixel's RGB color as:
\begin{align}
     \mathbf{C}(r) = \sum_{i=1}^{\npoints} T_i\alpha_i c_i,\text { where } T_i = \exp \left( -\sum_{j=1}^{i-1} \alpha_j \delta_j \right), \alpha_i = 1 - \exp(-\sigma_i \delta_i).
\end{align}

\noi Here,  $T_i$ denotes the transmittance accumulated up to a point $\point_i$, and $\alpha_i$ denotes the alpha value of the sampled point. $\npoints$ is the number of points sampled along the ray, and $\delta$ is the distance between neighboring sample points.
Setting $w_i = T_i\alpha_i$ gives
\begin{align}\label{eq:render color}
     \mathbf{C}(r) = \sum_{i=1}^{\npoints} w_i c_i,
\end{align}
where $w_i$ is the weight for the $i$-th sample along the ray. It is noted that the weight value $w_i$ increases as the sampled point $\point_i$ gets closer to the surface.
Similar to NeuS~\cite{wang2021neus}, we compute the density $\sigma$ from the SDF as
\begin{equation}
            \sigma(\point) = \max\left(\frac{-\frac{d\Phi_s\left(\sdf(\point(t))\right)}{dt}}{\Phi_s\left(\sdf(\point(t))\right)}, 0\right),
\end{equation} 
where $\Phi_s(x) = \left(1 + e^{-sx}\right)^{-1}$ is a sigmoid function, $s^{-1}$ is the trainable standard deviation which approaches $0$ as training converges and $t$ is the distance between $o$ and $\point$ as given in ~\Cref{eq:ray_eq}.

While density-based volume rendering methods such as NeRF \cite{mildenhall2021nerf}  can achieve high-quality novel view synthesis, they fail to capture accurate surface details due to their lack of distinct definition of a surface. NeuS~\cite{wang2021neus} showed that replacing the density $\sigma$ at a point $\point$ by its SDF value can significantly enhance the 3D reconstruction accuracy of neural-based representations.

Building on the recent success of NeuS~\cite{wang2021neus} for 3D shape representation, we choose to represent the object's shape using its SDF. The SDF $\sdf(\point)$ of the shape at a point $\point \in \rthree$  is defined as:
\begin{equation}
     \sdf : \rthree \mapsto \real, \quad \sdf(\point) = s(\point) \cdot d(\point),
\end{equation}
\noi where $d(\point)$ is the Euclidean distance from $\point$ to its nearest point on the surface, and $s(\point)$ is the sign function that takes the value of $-1$ if $\point$ is inside the surface and $+1$ if it is outside.  We parametrize the SDF function using an  MLP \cite{park2019deepsdf} $f_\networkparams$, referred to as the SDF network,  of the form:
\begin{equation}
    SDF, \hat{z} = f_\networkparams(\gamma(\point)),
\end{equation}

\noi where $ SDF\in\real$, $\networkparams$ are the learnable parameters of the MLP and $\gamma$ corresponds to positional encoding \cite{mildenhall2021nerf, tancik2020fourier} that maps $\point$ to a higher dimensional space. The SDF network also outputs a feature vector $\hat{z} \in\real^{256}$ that characterizes the geometry of the 3D shape.

In addition to predicting the SDF value, we use another MLP $g_\networkparams$, hereinafter referred to as the Neural Renderer network, to predict the RGB color value of the point $\point$ as viewed from the viewing direction $\viewdir$. It is represented as $g_\networkparams$:
\begin{equation}
    \hat{\mathbf{c}} = g_\networkparams(\point, \viewdir, \hat{\normalvec}, \hat{z}).
    \label{eq:color}
\end{equation}

\noi 
The final RGB color at $\point$ as viewed from  $\viewdir$  is then estimated using \Cref{eq:render color}. Note that we also condition the neural renderer network on the normal field $\hat{\normalvec}$ derived from the gradient of the estimated SDF, \ie $\hat{\normalvec}(\point) = \nabla_\point \sdf(\point)$. 

\subsection{Depth-Normal consistency-based supervision}
\label{sec:depth_normal_consistency} 
As will be demonstrated in \Cref{sec:results}, training neural surface-based methods such as NeuS~\cite{wang2021neus} and its variants requires a dense multi-view stereo setup, \ie one needs to capture a large number of images from multiple viewpoints and with a significant overlap between the images. Still, these methods fail to capture surface details and can result in large reconstruction errors. These errors become very significant when we reduce the number of images or reduce the overlap between the images. We argue that this is mainly because the supervisory signals used in these methods, whether they are based on photometric loss, chamfer distance-based geometric loss, or depth- or 3D point-based losses, are of 0-order, \ie they are a property of isolated points and thus do not capture the shape of the local geometry. 

\subsubsection{Normal-based supervision} 
\label{sec:depth_normal_loss}
We propose to use, as supervisory signals,  the first-order differential properties of surfaces, \ie surface normals. By explicitly enforcing the gradient of the estimated signed distance field to align with the normals of the target surface, one can achieve highly accurate 3D reconstruction even in scenarios where only as few as two images of the object are available. Unlike points, normals are the property of local patches around a point. Their benefits and ability to capture local surface details have already been demonstrated in photometric stereo-based methods. 

We measure the depth-normal consistency loss between ground-truth normals $\normalvec$  and rendered normals $\hat{\normalvec}$ as
\begin{align}
    \mathcal{L}_{dnc} = \frac{1}{m} \sum_{i=1}^{m} \lVert \mathbf{T}_{c\rightarrow w}(\normalvec_i) - \hat{\normalvec}_i \rVert_2,
    \label{eq:normal_loss}
\end{align}
where $m$ is the number of pixels sampled from the image of the current view under consideration, and $\mathbf{T}_{c\rightarrow w}$ for the view transforms the normals from camera coordinates to world coordinates.

Next, we describe how to compute the ground-truth normal $\normalvec$ and rendered normal $\hat{\normalvec}$ in Sections \ref{sec:normal_from_depth} and \ref{sec:surface_extraction}, respectively.

\subsubsection{Estimating normals from depth}
\label{sec:normal_from_depth}

We estimate the ground truth normals $\normalvec$ directly from the input RGB images without relying on auxiliary input or on photometric stereo. Given $\nimages$ multi-view RGB images captured around the object of interest, we use the pre-trained off-the-shelf depth estimator, Depth Anything \cite{depthanything}, to estimate relative depth image $\depthmap$ for each image. Depth Anything is a monocular depth estimator, and thus, it returns relative depth values in the Normalized Device Coordinates (NDC), with values in the $[0, 1]$ range. Note that we do not have to worry about the uniform scaling of the estimated depth images since surface normals do not depend on the scale, thus they are the same irrespective of whether we consider relative or metric depth. Therefore, we first lift each pixel $i$ and its corresponding depth value to a 3D point $\point_i$, resulting in a 3D point cloud. We then estimate the normal at each 3D point by fitting a plane to the neighborhood of that point. We collect all the points within a certain radius and perform PCA. The first two leading eigenvectors $v_{i,1}$ and $v_{i,2}$ define a tangent plane to the object's surface at $\point_i$. The normal vector $\normalvec_i$ to the surface of the object at $\point_i$ and thus for pixel $i$ is obtained by taking the cross product of the two tangent vectors $v_{i,1}$ and $v_{i,2}$.

\subsubsection{Accurate normal estimation via surface point localization}
\label{sec:surface_extraction}

We obtain the rendered normals from the gradient of the estimated SDF function. To do so, we need to derive the surface point in a differentiable manner to enable end-to-end training. Classically, differentiable ray tracing \cite{yariv2020multiview} has been used to find the intersection point of a ray $\ray$ with the surface of the object. However, this requires repeatedly querying the SDF network to find the zero-level set. Instead, Mono-SDF \cite{Yu2022MonoSDF} and NeuRIS \cite{wang2022neuris} use alpha compositing to estimate the depth and surface normals. But, this introduces inherent errors in the reconstructed surface as discussed in \cite{wang2021neus}. Thus, in this paper, we approximate the zero-level set by taking advantage of the SDF values predicted by the SDF network at each point sampled along the ray.

The idea is to find two points $\point(t_k)$ and $\point(t_{k + 1})$ along the ray $\ray$ such that the SDF at the first point is positive and negative at the second point. We only consider the first ray-surface intersection point, ignoring the subsequent intersections. 
Mathematically, it is defined as 
\begin{equation}
     k = \arg\min_i \{ t_i ~ | ~ \sdf(\point(t_i)) > 0 ~\text{ and }~ \sdf(\point(t_{i + 1})) < 0 \}.
     \label{eq:intersection}
 \end{equation}

\noi This way, we are guaranteed that the surface is localized in between these two points. The minimization over $i$ in \Cref{eq:intersection} ensures that we are finding the smallest possible index for $t$ such that the first ray-object intersection occurs between $t_k$ and $t_{k+1}$. We now know that the surface point along the ray $\ray$ is localized between the points $\point(t_i)$ and $\point(t_{i + 1})$. Thus,  we estimate the intersection point $\hat{\point}$ by using linear interpolation as:
\begin{equation}
    \hat{\point} = 
\begin{aligned}
\point(\bar{t}) \text{ such that } \bar{t} = \frac{\sdf(\point(t_k))t_{k+1} - \sdf(\point(t_{k + 1}))t_k}{ \sdf(\point(t_k)) - \sdf(\point(t_{k + 1}))} 
    \end{aligned}.
\end{equation}

\noi With this formulation, the smaller the distance between $t_k$ and $t_{k+1}$ is, the better the accuracy of the surface point localization will be. 

Thus, we take advantage of hierarchical sampling. We first coarsely sample $n=16$ equidistant points along a ray and evaluate the importance of each sample based on its weight value $w_i$. Regions of high importance, which correspond to points near the surface, are subdivided into smaller regions, and additional samples are placed within these intervals.   This process is repeated multiple times, $16$ times in our implementation, so we sample the points at fine intervals,  providing accurate surface localization. Once  the surface point $\hat{\point}$ is localized, we can compute the unit normal vector $\hat{\normalvec}$ to the surface at $\hat{\point}$  in world coordinates as:
\begin{equation}
    \hat{\normalvec}(\hat{\point}) =  \frac{\nabla_\point \sdf(\hat{\point})}{\|\nabla_\point \sdf(\hat{\point})\|_2},
\end{equation}

\noi which is the gradient of the SDF.
The above computation does not result in any computation overheads or increase in the training time.

\subsubsection{Total training loss}

We train the proposed network in a fully self-supervised manner using a loss for a particular view:
\begin{equation}\label{eq:total loss}
\mathcal{L} = \mathcal{L}_{\text{color}} + \alpha\mathcal{L}_{\text{eik}} + \beta\mathcal{L}_{\text{dnc}}.
\end{equation}

\noi It is a weighted sum of three terms: a photometric loss term $\mathcal{L}_{\text{color}} $, an  Eikonal loss term $\alpha\mathcal{L}_{\text{eik}}$, and the depth-normal consistency loss $\mathcal{L}_{\text{dnc}}$. The latter is given by~\Cref{eq:normal_loss}.

\paragraph{Photometric Loss} Also referred to as the reconstruction loss, it measures, in terms of the $L_1$ metric, the discrepancy between the ground-truth pixel color $C_{i}$ and the rendered color $\hat{C}_{i}$ at the same pixel:
\begin{align}
    \mathcal{L}_{color} = \frac{1}{m} \sum_{i=1}^{m} \lVert C_i - \hat{C}_i \rVert_{1}.
\end{align}

\paragraph{Eikonal Loss}
To encourage the learned SDF to be a valid distance field, we add an  Eikonal regularization term \cite{gropp2020implicit} of the form:
\begin{equation}
    \mathcal{L}_{eik} = \frac{1}{n_B} \sum_{\point \in B} \left( \left\| \nabla_\point \sdf(\point) \right\|_2 - 1 \right)^2,
\end{equation}
where $B$ represents the set of $m \times n_p$ points sampled along each ray in the image of the current view, and $n_B$ denotes the number of all sampled points in $B$. Note that Eikonal loss is applied only on the SDF value, not the feature vector $\hat{z}$.

\begin{figure}[tb]
    \centering
    \resizebox{0.75\textwidth}{!}{
    \begin{tabularx}{\textwidth}{>{\centering\arraybackslash}X>{\centering\arraybackslash}X>{\centering\arraybackslash}X>{\centering\arraybackslash}X>{\centering\arraybackslash}X>{\centering\arraybackslash}X>{\centering\arraybackslash}X>{\centering\arraybackslash}X}
    \footnotesize\textbf{\textcolor{green}{Ours}} & \footnotesize\textbf{RNb-NeuS} & \footnotesize\textbf{Kaya23} & \footnotesize\textbf{PS-NeRF} & \footnotesize\textbf{Kaya22} & \footnotesize\textbf{\textcolor{green}{NeuS}} & \footnotesize\textbf{Li19} & \footnotesize\textbf{\textcolor{green}{Park16}} \\

        \multicolumn{8}{c}{
            \includegraphics[width=\textwidth]{images/normal_mae/bear_normal_mae.png}
        } \\
        \multicolumn{8}{c}{\footnotesize\textbf{(a) Bear}} \\

        \multicolumn{8}{c}{
            \includegraphics[width=\textwidth]{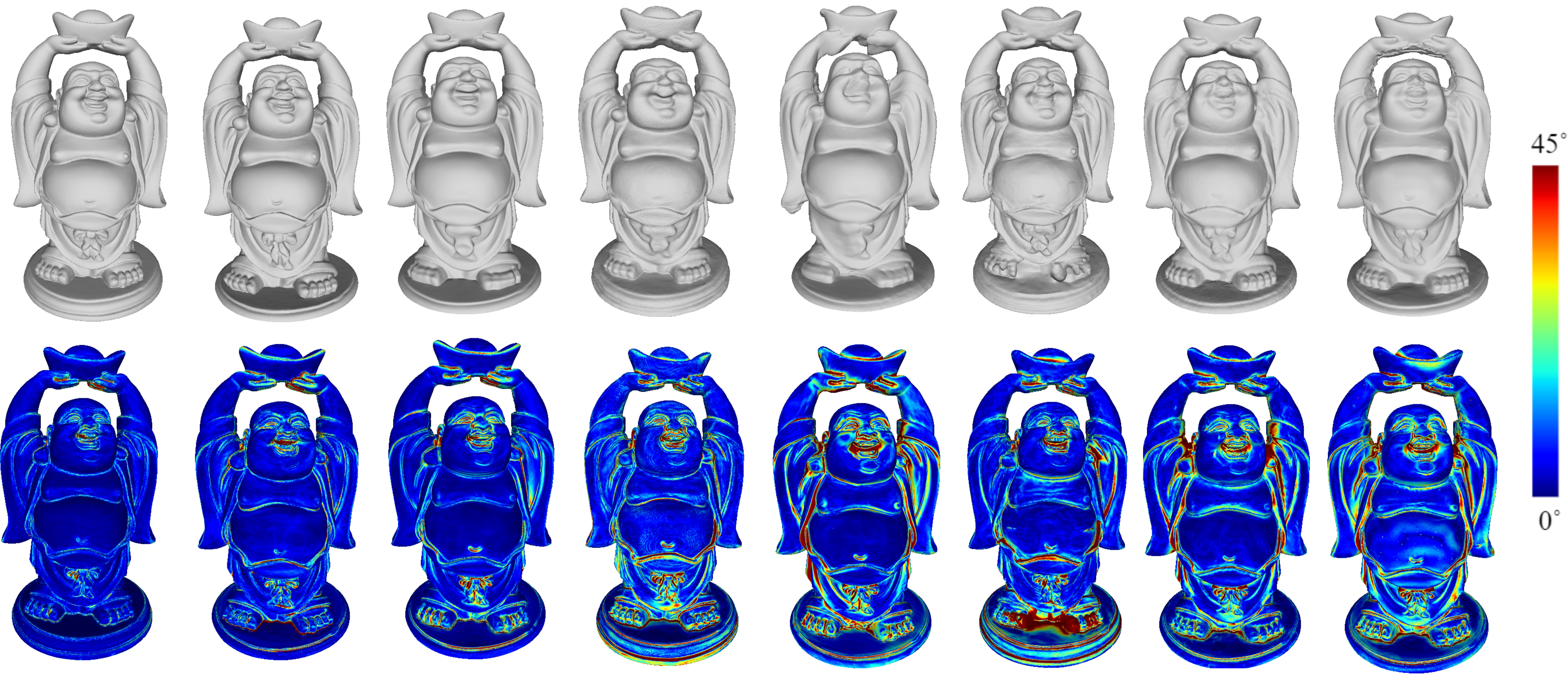}
        } \\
        \multicolumn{8}{c}{\footnotesize\textbf{(b) Buddha}} \\

        \multicolumn{8}{c}{
            \includegraphics[width=\textwidth]{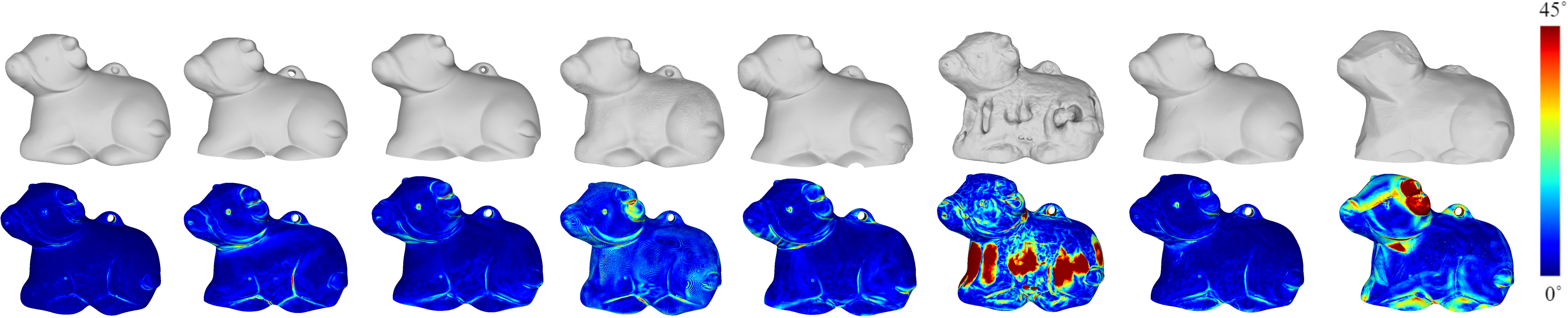}
        } \\
        \multicolumn{8}{c}{\footnotesize\textbf{(c) Cow}} \\

        \multicolumn{8}{c}{
            \includegraphics[width=\textwidth]{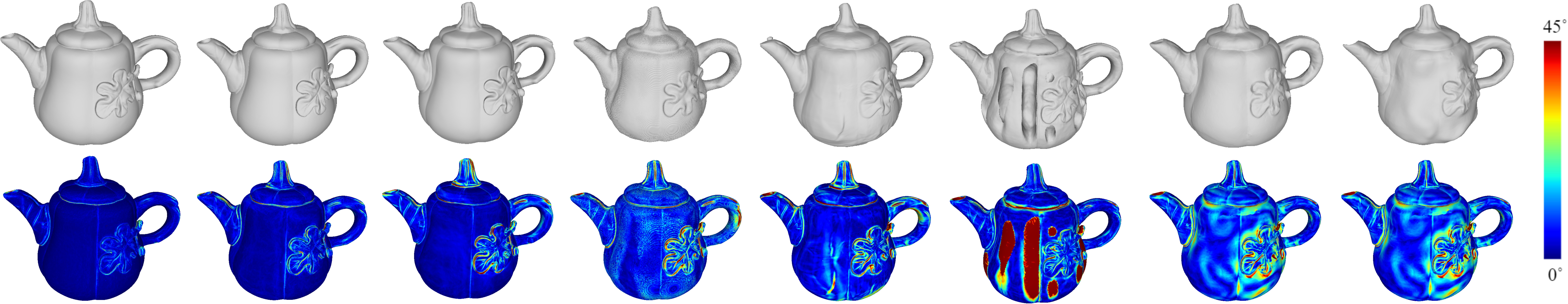}
        } \\
        \multicolumn{8}{c}{\footnotesize\textbf{(d) Pot2}} \\

        \multicolumn{8}{c}{
            \includegraphics[width=\textwidth]{images/normal_mae/reading_normal_mae.png}
        } \\
        \multicolumn{8}{c}{\footnotesize\textbf{(e) Reading}} 

    \end{tabularx}}
    \caption{Reconstructed 3D meshes and corresponding angular errors of three objects from the DiLiGenT-MV benchmark using $20$ views per object. The methods highlighted in \textcolor{green}{green} are MVS-based, while the others are MVPS-based.}
    \label{fig:plots_normal_mae}
\end{figure}



\begin{figure}[tb]
    \centering
    \begin{tabularx}{\textwidth}{>{\centering\arraybackslash}X
    >{\centering\arraybackslash}X>{\centering\arraybackslash}X
    >{\centering\arraybackslash}X>{\centering\arraybackslash}X
    >{\centering\arraybackslash}X>{\centering\arraybackslash}X
    >{\centering\arraybackslash}X>{\centering\arraybackslash}X
    >{\centering\arraybackslash}X}
    \footnotesize\textbf{Input image} & \footnotesize\textbf{\textcolor{green}{Park16}} & \footnotesize\textbf{Li19} & \footnotesize\textbf{\textcolor{green}{NeuS}} & \footnotesize\textbf{Kaya22} & \footnotesize\textbf{PS-NeRF} & \footnotesize\textbf{Kaya23} & \footnotesize\textbf{RNb-NeuS} & \footnotesize\textbf{\textcolor{green}{Ours}} & \footnotesize\textbf{GT}\\
    \multicolumn{10}{c}{
    \includegraphics[width=\textwidth]{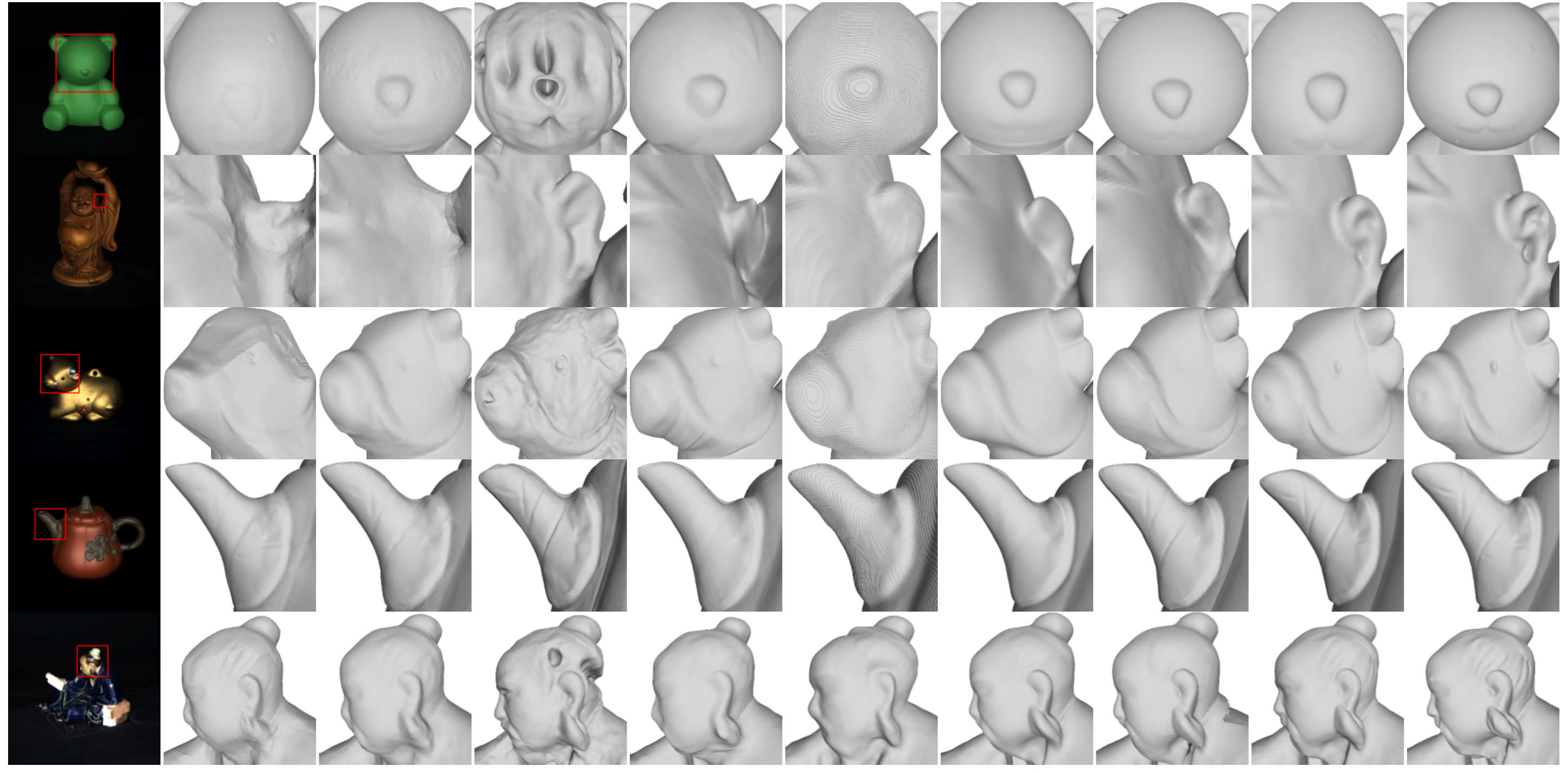}}
    \end{tabularx}
    \caption{Qualitative comparison, on the DiLiGenT-MV dataset, between our results and those obtained using state-of-the-art methods, with a zoom-in on parts of the meshes representing fine details. We use $20$ views per object for all methods. The methods highlighted in \textcolor{green}{green} are MVS-based, while the others are MVPS-based. }
    \label{fig:mesh_comp_all}
\end{figure}

\subsection{Implementation details}
\label{sec:implementation_details}
Similar to IDR~\cite{yariv2020multiview}, we use a neural network composed of two MLPs. The first is an SDF MLP used to calculate the SDF value of the desired surface given a  3D point $\point$ as input. It comprises $8$ layers of $256$ neurons each. It outputs the SDF value and a $256$-D feature vector $\hat{z}$. A second four-layer linear MLP is a neural renderer to compute the photometric loss. It also has $256$ neurons per layer and takes the feature vector $\hat{z}$,  the 3D point $\point$, a viewing direction $\viewdir$, and the gradient of the SDF $\nabla_\point \sdf(\point)$ to predict the RGB color value. We ensure that the target object of interest lies within the unit sphere as assumed by NeuS~\cite{wang2021neus}.


We train the entire network for $300$ epochs, which takes approximately $6$ hours on a single Nvidia A100 GPU. We use $512$ rays per image and  $256$ sample points per ray in all our experiments. We set the weights $\alpha$ and $\beta$  of \Cref{eq:total loss} to $0.1$ and $0.5$, respectively. After training, we use the Marching Cubes algorithm~\cite{Lorensen1987Marching} to extract a mesh from the SDF.
\section{Results}
\label{sec:results}

\subsection{Experimental setup}
\label{sec:experimental_setup}
 
We evaluated our approach on three datasets: 
\begin{enumerate}

\item The DiLiGenT-MV benchmark \cite{diligentMV} is a synthetic dataset that contains five objects \textit{(Bear, Buddha, Cow, Pot2}, and \textit{Reading)} with diverse shapes and materials. Each object has been imaged from  $20$ different views, with $96$ images per view captured under varying light directions and intensities.

\item The BlendedMVS dataset \cite{yao2020blendedmvs} is a real-world multi-view stereo data sets. In this dataset, each scene is captured from multiple viewpoints, with   $24$ to $143$ RGB images per scene. Each RGB image has a resolution of $576 \times 768$. We use nine challenging scenes in our experiments, namely \textit{(Bear, Dog, Durian, Jade, Man, Clock, Stone, Fountain, and Gundam)}.

\item The DTU dataset \cite{jensen2014large}  contains $49$ to $64$ RGB images at a resolution of $1200 \times 1600$ for each object scan with camera intrinsics and poses. Different approaches differ in the choice of input views when used for performance evaluation. In SparseNeuS setting~\cite{long2022sparseneus},
views $23, 24$, and $33$ of each scan are used. In this setting, the selected views are very close to each other, and there is a large overlap between them. In PixelNeRF~\cite{yu2021pixelnerf} setting,  
views $22, 25$, and $28$ of each scan are used. In this setup, the views are scattered, and there is little overlap between them. We conduct experiments on both settings.

\end{enumerate}
We use two metrics to quantitatively evaluate the performance of the proposed method, namely the Chamfer Distance (CD) and the Normal Mean Angular Error (MAE), which measure the quality of the 3D reconstruction results in terms of the geometric and surface normal accuracy. The Chamfer distance, measured in $mm$, quantifies the geometric similarity between the reconstructed and ground-truth meshes. The MAE, measured in degrees, measures the deviation of the estimated surface normals from the ground truth normals provided in the DiLiGenT-MV dataset.

\subsection{Comparisons with existing methods}
\label{sec:comparison}

\subsubsection{Evaluation on the DiLiGenT-MV dataset~\cite{diligentMV}}
We first evaluate our method on the 
DiLiGenT-MV dataset~\cite{diligentMV} and compare its performance to state-of-the-art MVS methods such as NeuS~\cite{wang2021neus}, Park16~\cite{park2016robust}, and MVPS methods such as Li19~\cite{li2020multi}, NeuS~\cite{wang2021neus}, Kaya22~\cite{kaya2022uncertainty}, Kaya23~\cite{kaya2023neural}, PS-NeRF~\cite{yang2022ps}, RNb-NeuS~\cite{Brument23} and SuperNormal \cite{supernormal2024cao}. In this experiment, we use all the $20$ views per model.

~\Cref{fig:plots_normal_mae} shows, for each example in the DiLiGenT-MV dataset~\cite{diligentMV},  the reconstructed 3D mesh and a surface plot of the corresponding reconstruction error measured in terms of angular error.
Our method, which is MVS-based, reconstructs surfaces of better quality and can recover fine details, outperforming both MVS methods such as NeuS and MVPS methods such as PS-NeRF and RNb-NeuS. 
 NeuS \cite{wang2021neus}, which is an MVS method, achieves reasonable shape reconstructions. However, the reconstructed surfaces contain significant noise and have noticeable holes or missing regions, particularly in some parts of the \textit{Bear} mesh.
 The plotted normal mean angular errors confirm this observation since we can see many red areas that correspond to a high angular error between the vertex normals of the reconstructed and the ground-truth meshes.
On the other hand, PS-NeRF~\cite{yang2022ps}, which is an MVPS method, recovers better the fine details of the overall shape and achieves superior reconstruction results than NeuS. 

~\Cref{fig:mesh_comp_all} provides a zoom-in on the reconstructed meshes using our method and compares them to the reconstructions obtained using state-of-the-art methods for all objects in the DiLiGenT-MV dataset. We can observe from the figure that introducing surface normals significantly improves the reconstruction quality and helps recover fine geometric details of the surface. Our method is the only one that achieves high-fidelity reconstruction on the ears of Buddha, the eyes and lasso on the Cow's face, the snout of Pot2, and the hairs of Reading. 

The MVPS method with the best performance is RNb-NeuS. However, we can see that it fails to capture surface details; see the face and book of the \emph{Reading} example (last row). 

 
\begin{table}[t]
\caption{Quantitative comparison, in terms of CD and MAE, of our method to multi-view stereo (MS) and multi-view photometric stereo (MVPS)  methods on the  DiLiGenT-MV dataset.
Best results are highlighted as \colorbox{colorFst}{1st}, \colorbox{colorSnd}{2nd}, and \colorbox{colorTrd}{3rd}.}
\label{table: 20 views}
\resizebox{0.8\textwidth}{!}{
\begin{tabular}{@{}l|*{5}{c}|*{5}{c}@{}}
    \toprule
    & \multicolumn{5}{c|}{\textbf{Chamfer Dist}($\downarrow$) } &  \multicolumn{5}{c}{\textbf{Normal MAE}$(\downarrow$)} \\
    \cline{2-11}
    Method 
    & \emph{Bear}  & \emph{Buddha}   & \emph{Cow}   & \emph{Pot2}   & \emph{Reading} 
     & \emph{Bear}  & \emph{Buddha}   & \emph{Cow}   & \emph{Pot2}   & \emph{Reading} 
    \\
    \hline
    NeuS~\cite{wang2021neus}
    & 35.02 & 10.64 & 27.07 & 34.59 & \colorbox{colorTrd}{14.88}
    & 20.25  & 11.72  & 18.66 & 19.02  & 16.49 
    \\
    \hline
    
    Park16~\cite{park2016robust}
    &  19.58  & 11.77  & \colorbox{colorSnd}{9.25}  & 24.82  &  22.62 
    & 12.78  & 14.68  & 13.21  & 15.53  & 12.92 
    \\
    Li19~\cite{li2020multi}
    &  \colorbox{colorTrd}{8.91} &  13.29 &  14.01 & \colorbox{colorTrd}{7.40}  &  24.78 
    & \colorbox{colorTrd}{4.39}  & \colorbox{colorTrd}{11.45}  & 4.14  & 6.70  & 8.73 
    \\
    
    Kaya22~\cite{kaya2022uncertainty}
    & 381.66 & 416.77 & 311.52 & 502.83 & 346.05 
    & 5.90 & 20.04 & 6.04 & 12.68 & 8.21
    \\
    PS-NeRF~\cite{yang2022ps}
    & \colorbox{colorSnd}{8.65} & \colorbox{colorTrd}{8.61} & \colorbox{colorTrd}{10.21} & \colorbox{colorSnd}{6.11} & \colorbox{colorSnd}{12.35}
    & 5.48  & 11.7  & 5.46 & 7.65  & 9.13 
    \\
    Kaya23~\cite{kaya2023neural}
    & 330.37 & 198.96 & 215.18 & 366.30 & 292.87
    & 4.83 & 12.06 & \colorbox{colorTrd}{3.75} & 8.06 & \colorbox{colorTrd}{7.06} 
    \\
    RNb-NeuS~\cite{Brument23}
    & 38.19 & \colorbox{colorSnd}{7.69} & 42.78 & 7.68 & 15.57
    & \colorbox{colorSnd}{2.70} & \colorbox{colorSnd}{8.17}  & \colorbox{colorSnd}{3.61} & \colorbox{colorTrd}{4.11}  & \colorbox{colorSnd}{6.18}  
    \\
    SuperNormal~\cite{supernormal2024cao}
    & 27.79 & 16.28 & 19.74 & 10.56 & 18.33
    & 7.14  & 12.98  & 5.22 & \colorbox{colorSnd}{3.32}  & 8.35  
    \\
    \hline
    Ours
    & \colorbox{colorFst}{8.62} & \colorbox{colorFst}{4.82} & \colorbox{colorFst}{8.99} & \colorbox{colorFst}{5.54} & \colorbox{colorFst}{7.60}
    & \colorbox{colorFst}{0.87}  & \colorbox{colorFst}{3.49}  & \colorbox{colorFst}{1.09}  & \colorbox{colorFst}{1.85}  & \colorbox{colorFst}{4.04}
    \\
    \bottomrule
\end{tabular}
}
\end{table}


~\Cref{table: 20 views} shows the quantitative evaluation and comparison of current methods with ours in terms of CD and normal MAE. In this experiment, we use $20$ views per object. However, MVPS methods use multiple images per view, each captured under different lighting conditions. Our method uses only one image per view. The table shows that our method significantly outperforms current state-of-the-art methods on all five objects by a large margin for both the Chamfer distance and normal MAE metric.
A similar trend can be observed for the normal MAE, where our method attains the lowest angular error for all five objects in the dataset. 

\begin{figure}[!ht]
\centering
    \scalebox{0.75}{
    \begin{tikzpicture}[spy using outlines={circle,size=2.5cm,connect spies,every spy on node/.append style={thick}}]
    \node{\includegraphics[width=\textwidth]{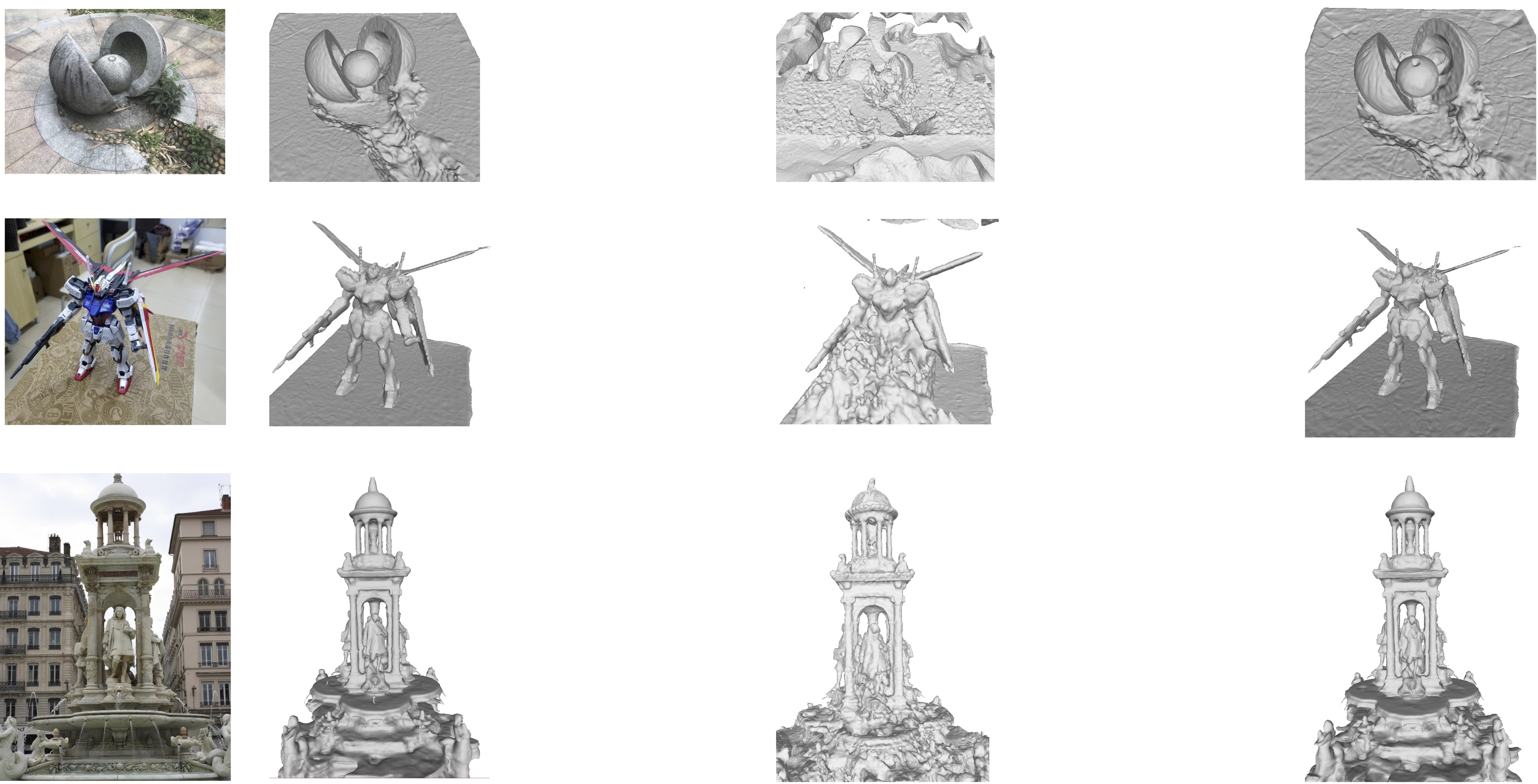}};
    \spy[color=red,magnification=2] on (-3.7, 3.15) in node [left] at (-0.1, 3);
    \spy[color=red,magnification=2] on (1.15, 3) in node [left] at (4.7, 3);
    \spy[color=red,magnification=2] on (5.8, 3.) in node [left] at (9.7, 3);        
    \spy[color=green,magnification=3] on (-3.55, 1) in node [left] at (-0.1, 0.25);   
    \spy[color=green,magnification=3] on (1.15, 1) in node [left] at (4.7, 0.25); 
    \spy[color=green,magnification=3] on (5.85, 1) in node [left] at (9.7, 0.25);          
    \spy[color=cyan,magnification=4]  on (-3.55, -2) in node [left] at (-0.1, -2.5);
    \spy[color=cyan,magnification=4]  on (0.965, -2.1) in node [left] at (4.7, -2.5);
    \spy[color=cyan,magnification=4]  on (5.8, -2) in node [left] at (9.7, -2.5);    
    \node[rotate=90] at (-8., 2.8) {\footnotesize\textbf{Stone}};   
    \node[rotate=90] at (-8., 0.6) {\footnotesize\textbf{Gundam}};   
    \node[rotate=90] at (-8., -2.25) {\footnotesize\textbf{Fountain}};
    \node at (-6.0, -4.2) {\small \textbf{Input image}};
    \node at (-3.5, -4.2) {\small \textbf{Ours}};
    \node at (1.8, -4.2) {\small \textbf{NeuS2~\cite{wang2023neus2}}};
    \node at (7, -4.2) {\small \textbf{NeuS~\cite{wang2021neus}}};   
    \end{tikzpicture}}
    \caption{Qualitative comparison of surface reconstruction results on the BlendedMVS dataset.}
    \label{fig:bmvs_mesh_comp}
\end{figure}

\subsubsection{Evaluation on BlendedMVS dataset~\cite{yao2020blendedmvs}} 
Since BlendedMVS is an MVS dataset, we only consider MVS methods such as NeuS \cite{wang2021neus} and NeuS2~\cite{wang2023neus2}. We use the Chamfer distance for the quantitative evaluation and perform two experiments. In the first one, we use all the views available for each object. In the second experiment, we only use $8$ views.  ~\Cref{tab:cmp_bmvs_all} summarizes the results. In most cases, our method performs better than other methods using all views, while it outperforms all other methods when we only have sparse $8$ views. This shows that our method is particularly effective in sparse-view scenarios.

\begin{table}[t]
\centering
\caption{Quantitative evaluation, in terms of CD ($\downarrow$), of the reconstruction quality on the BlendedMVS dataset and comparison with the state-of-the-art MVS methods. Best results are highlighted as \colorbox{colorFst}{1st} and \colorbox{colorSnd}{2nd}.}\label{tab:cmp_bmvs_all}\label{tab:cmp_bmvs_8views}
\resizebox{0.8\textwidth}{!}{
\begin{tabular}{@{}c|ccc|ccc@{}}
\toprule
      & \multicolumn{3}{c|}{\textbf{Using all views}} & \multicolumn{3}{c}{\textbf{Using $8$ views}} \\
\cline{2-7}
 & Ours & NeuS\cite{wang2021neus} & NeuS2\cite{wang2023neus2}  & Ours & NeuS\cite{wang2021neus} & NeuS2\cite{wang2023neus2} \\ \hline
Bear      & \colorbox{colorFst}{2.27} & \colorbox{colorSnd}{3.13} & 3.25 & \colorbox{colorFst}{3.56} & \colorbox{colorSnd}{3.91} & 4.34 \\
Clock     & \colorbox{colorFst}{2.54} & \colorbox{colorSnd}{2.77} & 2.92 & \colorbox{colorFst}{5.78} & \colorbox{colorSnd}{7.09} & 7.41\\
Dog       & \colorbox{colorFst}{2.48} & \colorbox{colorSnd}{2.76} & 3.06 & \colorbox{colorFst}{4.64} &\colorbox{colorSnd}{ 5.16} & 6.02\\
Durian    & \colorbox{colorFst}{2.18} & \colorbox{colorSnd}{3.17} & 3.34  & \colorbox{colorFst}{6.67} & \colorbox{colorSnd}{7.12} & 7.66 \\
Jade      & \colorbox{colorFst}{3.47} & 4.27 & \colorbox{colorSnd}{4.06} & \colorbox{colorFst}{6.29} & 8.50 & \colorbox{colorSnd}{8.05}\\ 
Man       & \colorbox{colorFst}{2.07} & 2.26 & \colorbox{colorSnd}{2.18} & \colorbox{colorFst}{3.45} & 4.08 & \colorbox{colorSnd}{3.88}\\
Stone      & \colorbox{colorFst}{2.09} & \colorbox{colorSnd}{2.81} & 4.18 & \colorbox{colorFst}{3.12} & \colorbox{colorSnd}{3.68} & 5.16 \\
Fountain       & \colorbox{colorFst}{3.21} & \colorbox{colorSnd}{3.65} & 4.44  & \colorbox{colorFst}{5.54} & 7.61 & \colorbox{colorSnd}{5.49}\\
Gundam      & \colorbox{colorFst}{1.28} & \colorbox{colorSnd}{1.64} & 4.67  & \colorbox{colorFst}{2.23} & \colorbox{colorSnd}{3.11} & 5.27 \\
\hline
Mean      & \colorbox{colorFst}{2.39} & \colorbox{colorSnd}{2.94} & 3.57  & \colorbox{colorFst}{4.58} & \colorbox{colorSnd}{5.58} & 5.92 \\ 
\bottomrule
\end{tabular}
}
\end{table}

\begin{table*}[!ht]
    \centering
    \caption{The quantitative comparison results of Chamfer Distances (CD$\downarrow$) on the DTU dataset for sparse views (three) under large-overlap and little-overlap settings. Best results are highlighted as \colorbox{colorFst}{1st}, \colorbox{colorSnd}{2nd}, and \colorbox{colorTrd}{3rd}.}
    \resizebox{\textwidth}{!}{
        \begin{tabular}{l|*{15}{c}|c}
            \toprule
            \textbf{Scan ID} & \textbf{24} & \textbf{37} & \textbf{40} & \textbf{55} & \textbf{63} & \textbf{65} & \textbf{69} & \textbf{83} & \textbf{97} & \textbf{105} & \textbf{106} & \textbf{110} & \textbf{114} & \textbf{118} & \textbf{122} & \textbf{Mean} \\
            \midrule
            {COLMAP \cite{schonberger2016structure}} & \colorbox{colorTrd}{2.88} & \colorbox{colorTrd}{3.47} & \colorbox{colorSnd}{1.74} & 2.16 & 2.63 & 3.27 & 2.78 & 3.63 & 3.24 & 3.49 & 2.46 & \colorbox{colorTrd}{1.24} & 1.59 & 2.72 & 1.87 & 2.61\\
            \cmidrule(l{0.7em}r{0.7em}){1-17}
            {SparseNeuS \cite{long2022sparseneus}} & 4.81 & 5.56 & 5.81 & 2.68 & 3.30 & 3.88 & 2.39 & 2.91 & 3.08 & 2.33 & 2.64 & 3.12 & 1.74 & 3.55 & 2.31 & 3.34  \\
            {VolRecon \cite{ren2023volrecon}} & 3.05 & 4.45 & 3.36 & 3.09 & 2.78 & 3.68 & 3.01 & 2.87 & 3.07 & 2.55 & 3.07 & 2.77 & 1.59 & 3.44 & 2.51 & 3.02 \\
            \cmidrule(l{0.7em}r{0.7em}){1-17}
            NeuS \cite{wang2021neus} & 4.11 & 5.40 & 5.10 & 3.47 & 2.68 & \colorbox{colorTrd}{2.01} & 4.52 & 8.59 & 5.09 & 9.42 & 2.20 & 4.84 & 0.49 & 2.04 & 4.20 & 4.28 \\
            VolSDF \cite{yariv2021volume} & 4.07 & 4.87 & 3.75 & 2.61 & 5.37 & 4.97 & 6.88 & 3.33 & 5.57 & 2.34 & 3.15 & 5.07 & \colorbox{colorTrd}{1.20} & 5.28 & 5.41 & 4.26  \\
            MonoSDF \cite{Yu2022MonoSDF} & 3.47 & 3.61 & \colorbox{colorTrd}{2.10} & \colorbox{colorTrd}{1.05} & \colorbox{colorTrd}{2.37} & \colorbox{colorSnd}{1.38} & \colorbox{colorTrd}{1.41} & \colorbox{colorTrd}{1.85} & \colorbox{colorTrd}{1.74} & \colorbox{colorTrd}{1.10} & \colorbox{colorTrd}{1.46} & 2.28 & 1.25 & \colorbox{colorTrd}{1.44} & \colorbox{colorTrd}{1.45} & \colorbox{colorTrd}{1.86} \\
            NeuSurf \cite{huang2024neusurf} & \colorbox{colorFst}{1.35} & \colorbox{colorSnd}{3.25} & 2.50 & \colorbox{colorSnd}{0.80} & \colorbox{colorFst}{1.21} & 2.35 & \colorbox{colorSnd}{0.77} & \colorbox{colorSnd}{1.19} & \colorbox{colorSnd}{1.20} & \colorbox{colorSnd}{1.05} & \colorbox{colorSnd}{1.05} & \colorbox{colorSnd}{1.21} & \colorbox{colorFst}{0.41} & \colorbox{colorSnd}{0.80} & \colorbox{colorSnd}{1.08} & \colorbox{colorSnd}{1.35} \\
            Gaussian Surfels \cite{Dai2024GaussianSurfels} & 4.96 & 4.72 & 4.41 & 3.84 & 4.84 & 4.38 & 5.56 & 6.49 & 4.75 & 5.39 & 4.42 & 6.76 & 4.25 & 3.31 & 4.16 & 4.82\\
            \cmidrule(l{0.7em}r{0.7em}){1-17}
            Ours & \colorbox{colorSnd}{1.54} & \colorbox{colorFst}{2.01} & \colorbox{colorFst}{1.46} & \colorbox{colorFst}{0.74} & \colorbox{colorSnd}{2.14} & \colorbox{colorFst}{1.25} & \colorbox{colorFst}{0.75} & \colorbox{colorFst}{0.92} & \colorbox{colorFst}{1.13} & \colorbox{colorFst}{0.88} & \colorbox{colorFst}{0.78} & \colorbox{colorFst}{0.95} & \colorbox{colorSnd}{0.56} & \colorbox{colorFst}{0.70} & \colorbox{colorFst}{0.82} & \colorbox{colorFst}{1.11}\\
            \bottomrule
            \multicolumn{17}{c}{(a) Sparse Multi-view Stereo with Little-overlap (PixelNeRF Setting)}\\
            \multicolumn{17}{c}{}\\
            \toprule
            \textbf{Scan ID} & \textbf{24} & \textbf{37} & \textbf{40} & \textbf{55} & \textbf{63} & \textbf{65} & \textbf{69} & \textbf{83} & \textbf{97} & \textbf{105} & \textbf{106} & \textbf{110} & \textbf{114} & \textbf{118} & \textbf{122} & \textbf{Mean} \\
            \midrule
            {COLMAP \cite{schonberger2016structure}} & \colorbox{colorTrd}{0.90} & 2.89 & 1.63 & 1.08 & 2.18 & 1.94 & 1.61 & \colorbox{colorTrd}{1.30} & 2.34 & 1.28 & \colorbox{colorTrd}{1.10} & 1.42 & 0.76 & \colorbox{colorTrd}{1.17} & \colorbox{colorTrd}{1.14} & 1.52 \\
            \cmidrule(l{0.7em}r{0.7em}){1-17}
            {SparseNeuS \cite{long2022sparseneus}} & 2.17 & 3.29 & 2.74 & 1.67 & 2.69 & 2.42 & 1.58 & 1.86 & 1.94 & 1.35 & 1.50 & 1.45 & 0.98 & 1.86 & 1.87 & 1.96 \\
            {VolRecon \cite{ren2023volrecon}} & 1.20 & \colorbox{colorTrd}{2.59} & \colorbox{colorTrd}{1.56} & 1.08 & \colorbox{colorSnd}{1.43} & 1.92 & \colorbox{colorTrd}{1.11} & 1.48 & \colorbox{colorTrd}{1.42} & \colorbox{colorTrd}{1.05} & 1.19 & \colorbox{colorTrd}{1.38} & \colorbox{colorTrd}{0.74} & 1.23 & 1.27 & \colorbox{colorTrd}{1.38} \\
            \cmidrule(l{0.7em}r{0.7em}){1-17}
            NeuS \cite{wang2021neus} & 4.57 & 4.49 & 3.97 & 4.32 & 4.63 & 1.95 & 4.68 & 3.83 & 4.15 & 2.50 & 1.52 & 6.47 & 1.26 & 5.57 & 6.11 & 4.00  \\
            VolSDF \cite{yariv2021volume} & 4.03 & 4.21 & 6.12 & \colorbox{colorTrd}{0.91} & 8.24 & \colorbox{colorTrd}{1.73} & 2.74 & 1.82 & 5.14 & 3.09 & 2.08 & 4.81 & 0.60 & 3.51 & 2.18 & 3.41 \\
            MonoSDF \cite{Yu2022MonoSDF} & 2.85 & 3.91 & 2.26 & 1.22 & 3.37 & 1.95 & 1.95 & 5.53 & 5.77 & 1.10 & 5.99 & 2.28 & 0.65 & 2.65 & 2.44 & 2.93 \\
            NeuSurf \cite{huang2024neusurf} & \colorbox{colorSnd}{0.78} & \colorbox{colorSnd}{2.35} & \colorbox{colorSnd}{1.55} & \colorbox{colorSnd}{0.75} & \colorbox{colorFst}{1.04} & \colorbox{colorSnd}{1.68} & \colorbox{colorSnd}{0.60} & \colorbox{colorSnd}{1.14} & \colorbox{colorSnd}{0.98} & \colorbox{colorFst}{0.70} & \colorbox{colorSnd}{0.74} & \colorbox{colorFst}{0.49} & \colorbox{colorSnd}{0.39} & \colorbox{colorSnd}{0.75} & \colorbox{colorSnd}{0.86} & \colorbox{colorSnd}{0.99} \\
            Gaussian Surfels \cite{Dai2024GaussianSurfels} & 4.30 & 4.06 & 4.37 & 3.13 & 4.94 & 3.95 & 5.23 & 6.11 & 4.56 & 4.12 & 3.97 & 5.13 & 4.52 & 4.20 & 5.43 & 4.47\\
            \cmidrule(l{0.7em}r{0.7em}){1-17}
            Ours & \colorbox{colorFst}{0.61} & \colorbox{colorFst}{1.96} & \colorbox{colorFst}{1.33} & \colorbox{colorFst}{0.59} & \colorbox{colorTrd}{1.57} & \colorbox{colorFst}{1.31} & \colorbox{colorFst}{0.57} & \colorbox{colorFst}{0.72} & \colorbox{colorFst}{0.89} & \colorbox{colorSnd}{0.84} & \colorbox{colorFst}{0.54} & \colorbox{colorSnd}{0.60} & \colorbox{colorFst}{0.31} & \colorbox{colorFst}{0.49} & \colorbox{colorFst}{0.66} & \colorbox{colorFst}{0.87} \\
            \bottomrule
            \multicolumn{17}{c}{(b) Sparse Multi-view Stereo with Large-overlap (SparseNeuS Setting)} \\
        \end{tabular}
    }
    \label{tab:dtu_cd_results}
\end{table*}

~\Cref{fig:bmvs_mesh_comp}  compares NeuS~\cite{wang2021neus} and NeuS2~\cite{wang2023neus2} with our method on three objects from the BlendedMVS dataset. We observe that while NeuS2 \cite{wang2023neus2} is significantly faster than the other two methods, it is unable to recover high-quality meshes of real-world objects.
While NeuS \cite{wang2021neus} produces decent quality meshes, we can see some artifacts at the top and shells of \textit{Stone}  (first row) and around the shoulders of \textit{Gundam} (second row). Our method provides highly detailed and highly accurate meshes in regions such as the top of the ball in \textit{Stone}, the separation between the arm and arm shield in \textit{Gundam}, and the hand of the statue in \textit{Fountain}. 

\subsubsection{Evaluation on the DTU dataset~\cite{jensen2014large}}
\label{sec:dtu}
We also evaluate the performance of our method on the DTU dataset, which is an MVS dataset. In this experiment, we are particularly interested in demonstrating the performance of our method in sparse MVS when there is little overlap between the views and when there is a large overlap between the views. In both cases, we only use three views following the PixelNeRF~\cite{yu2021pixelnerf} and SparseNeuS~\cite{long2022sparseneus} settings. \Cref{tab:dtu_cd_results} reports the comparison, in terms of CD, of our method with $8$ state-of-the-art methods, including NeuSurf~\cite{huang2024neusurf}, which is based on neural surfaces, and Gaussian Surfels~\cite{Dai2024GaussianSurfels}, which is based on Gaussian splatting~\cite{kerbl3Dgaussians}. The former is specifically designed for sparse MVS settings. From this table, we can see that, on average, our method outperforms all the state-of-the-art techniques in the little and large overlap settings. We observe that in both settings, our method significantly outperforms Gaussian Surfels~\cite{Dai2024GaussianSurfels} in all scans and outperforms NeuSurf~\cite{huang2024neusurf} in $12$ scans out of $15$. In the supplementary material, we provide a visual comparison of the results of our method with the results of Gaussian Surfels~\cite{Dai2024GaussianSurfels}.

\begin{figure}
    \centering
    \includegraphics[width=\textwidth]{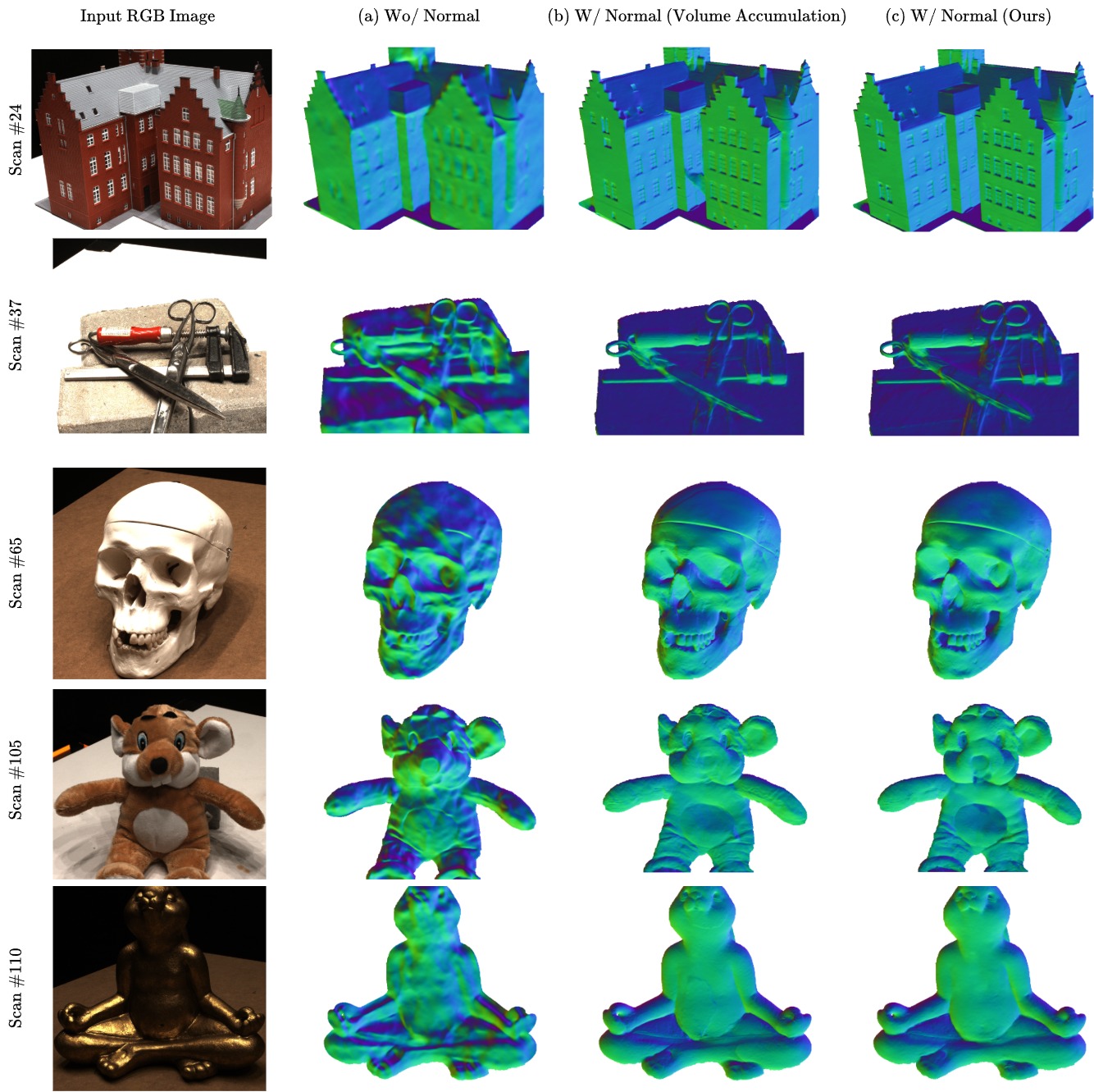}
    \caption{Rendered normal maps generated using our method for a given RGB image when (a) we do not use the normal loss ($\mathcal{L}_{\text{dnc}}$), (b) we add normal loss ($\mathcal{L}_{\text{dnc}}$) to the overall loss function, where normals are computed using volumetric accumulation along rays~\cite{wang2022neuris, Yu2022MonoSDF}, and in (c) we compute the normals using gradient of the SDF at the localized surface point.}
    \label{fig:svg-figure}
\end{figure}

~\Cref{fig:svg-figure} shows the rendered normal maps for various scenes generated using three different approaches: (a) without the normal loss ($\mathcal{L}_{\text{dnc}}$), (b) with the normal loss using volumetric accumulation along rays as done in~\cite{wang2022neuris, Yu2022MonoSDF}, and (c) with the normal loss using our proposed method of computing normals via the gradient of the SDF at localized surface points. As seen in the results, the method without normal loss (a) produces significantly noisier and less accurate normal maps, leading to poor representation of the surface details. Using normal loss with volumetric accumulation (b) improves the quality of the normal maps but still shows artifacts and inaccuracies, specifically in scan 24, scan 65 (upper part of the skull), and scan 110. In contrast, our method (c) produces the most accurate and clean normal maps, effectively capturing fine surface details and resulting in a better overall geometry representation.

\aptLtoX[graphic=no,type=html]{\begin{table}
  \centering
  \caption{Ablation studies on the DTU dataset, (a) effect of normal loss,
           (b) effect of different depth estimators, and (c) the effect of different normal extraction methods.}
  \label{tab:mergedAblations}
    \begin{tabular}{l c}
      \toprule
\multicolumn{2}{c}{(a)}\\
      \cline{1-2}
      \textbf{Loss} & \textbf{Mean CD} ($\downarrow$) \\
      \cline{1-2}
      Wo/ normal loss & 4.21 \\
      W/ normal loss & \textbf{0.84} \\
      \cline{1-2}
\multicolumn{2}{c}{(b)}\\
      \cline{1-2}
      \textbf{Depth estimator} & \textbf{Mean CD} ($\downarrow$) \\
      \cline{1-2}
      Omnidata~\cite{eftekhar2021omnidata} & 0.97 \\
      Depth Anything~\cite{depthanything} & \textbf{0.84} \\
      \cline{1-2}
\multicolumn{2}{c}{(c)}\\
      \cline{1-2}
      \textbf{Normal estimation method} & \textbf{Mean CD} ($\downarrow$) \\
      \cline{1-2}
      Vol. accum. along rays (Eqn.~\eqref{eq:1})  & 2.92 \\
      Gradient of SDF at surface pt. (Ours) & \textbf{0.84} \\
      \bottomrule
    \end{tabular}
\end{table}}{\begin{table*}[t]
  \centering
  \caption{Ablation studies on the DTU dataset, (a) effect of normal loss,
           (b) effect of different depth estimators, and (c) the effect of different normal extraction methods.}
  \label{tab:mergedAblations}
  \resizebox{\textwidth}{!}{
  \begin{subtable}[b]{0.3\textwidth}
    \centering
    \caption{}
    \label{tab:lossComparison}
    \begin{tabular}{l c}
      \toprule
      \textbf{Loss} & \textbf{Mean CD} ($\downarrow$) \\
      \midrule
      Wo/ normal loss & 4.21 \\
      W/ normal loss & \textbf{0.84} \\
      \bottomrule
    \end{tabular}
  \end{subtable}
    \hfill
  \begin{subtable}[b]{0.7\textwidth}
    \centering
    \caption{}
    \label{tab:depthComparison}
    \begin{tabular}{l c}
      \toprule
      \textbf{Depth estimator} & \textbf{Mean CD} ($\downarrow$) \\
      \midrule
      Omnidata~\cite{eftekhar2021omnidata} & 0.97 \\
      Depth Anything~\cite{depthanything} & \textbf{0.84} \\
      \bottomrule
    \end{tabular}
  \end{subtable}
  \hfill
   \begin{subtable}[b]{0.6\textwidth}
    \centering
    \caption{}
    \label{tab:normalComparison}
    \begin{tabular}{l c}
      \toprule
      \textbf{Normal estimation method} & \textbf{Mean CD} ($\downarrow$) \\
      \midrule
      Vol. accum. along rays (Eqn.~\eqref{eq:1})  & 2.92 \\
      Gradient of SDF at surface pt. (Ours) & \textbf{0.84} \\
      \bottomrule
    \end{tabular}
  \end{subtable}}
\end{table*}}

\aptLtoX[graphic=no,type=html]{\begin{table}
  \centering
  \caption{Ablation studies on the DTU dataset, (a) single MLP v/s full two-MLP model, (b) architectural choice for neural scene representation.}
  \label{tab:mergedAblations2}
    \begin{tabular}{l c}
      \toprule
\multicolumn{2}{c}{(a)}\\
      \cline{1-2}
      \textbf{Method} & \textbf{Mean CD} ($\downarrow$) \\
      \cline{1-2}
      Single MLP  & 1.57 \\
      Full two-MLP model (Ours) & \textbf{0.84} \\
      \cline{1-2}
\multicolumn{2}{c}{(b)}\\
      \cline{1-2}
      \textbf{Scene representation choice} & \textbf{Mean CD} ($\downarrow$) \\
      \cline{1-2}
      SDF Grid  & 1.05 \\
      SDF MLP (Ours) & \textbf{0.90} \\
      \bottomrule
    \end{tabular}
\end{table}}{\begin{table*}[t]
  \centering
  \caption{Ablation studies on the DTU dataset, (a) single MLP v/s full two-MLP model, (b) architectural choice for neural scene representation.}
  \label{tab:mergedAblations2}
  \resizebox{\textwidth}{!}{
  \begin{subtable}[b]{0.7\textwidth}
    \centering
    \caption{}
    \label{tab:single_network}
    \begin{tabular}{l c}
      \toprule
      \textbf{Method} & \textbf{Mean CD} ($\downarrow$) \\
      \midrule
      Single MLP  & 1.57 \\
      Full two-MLP model (Ours) & \textbf{0.84} \\
      \bottomrule
    \end{tabular}
  \end{subtable}
  \hfill
  \begin{subtable}[b]{0.7\textwidth}
    \centering
    \caption{}
    \label{tab:archiChoice}
    \begin{tabular}{l c}
      \toprule
      \textbf{Scene representation choice} & \textbf{Mean CD} ($\downarrow$) \\
      \midrule
      SDF Grid  & 1.05 \\
      SDF MLP (Ours) & \textbf{0.90} \\
      \bottomrule
    \end{tabular}
  \end{subtable}}
\end{table*}}

\aptLtoX[graphic=no,type=html]{\begin{table}[!ht]
\centering
\caption{Effect of using different numbers of layers in the SDF network on the reconstructed meshes when using the DTU dataset.}
\resizebox{0.2\textwidth}{!}{
\begin{tabular}{c|c}
\toprule
  \textbf{\# Layers} & \textbf{Mean CD ($\downarrow$)} \\ \midrule
5 & 1.12 \\
6 & 1.01 \\
7 & 0.93 \\
8 & \textbf{0.84} \\
9 & 0.88 \\
10 & 0.91 \\
\bottomrule
\end{tabular}}}{\begin{table}[!ht]
\centering
\caption{Effect of using different numbers of layers in the SDF network on the reconstructed meshes when using the DTU dataset.}
\resizebox{0.2\textwidth}{!}{
\begin{tabular}{c|c}
\hline
  \textbf{\# Layers} & \textbf{Mean CD ($\downarrow$)} \\ \hline
5 & 1.12 \\
6 & 1.01 \\
7 & 0.93 \\
8 & \textbf{0.84} \\
9 & 0.88 \\
10 & 0.91 \\
\hline
\end{tabular}}}

\label{tab:7}
\end{table}

\begin{figure}[!h] 
    \centering
    \setlength{\tabcolsep}{1pt} 
    \renewcommand{\arraystretch}{1.2} 
    \resizebox{0.8\textwidth}{!}{\begin{tabular}{@{}c@{}c@{}c@{}} 
        \begin{tikzpicture}[spy using outlines={rectangle, magnification=2, size=1.25cm, connect spies}]
        \node[anchor=south west,inner sep=0] (image) at (0,0) {\includegraphics[width=0.33\linewidth]{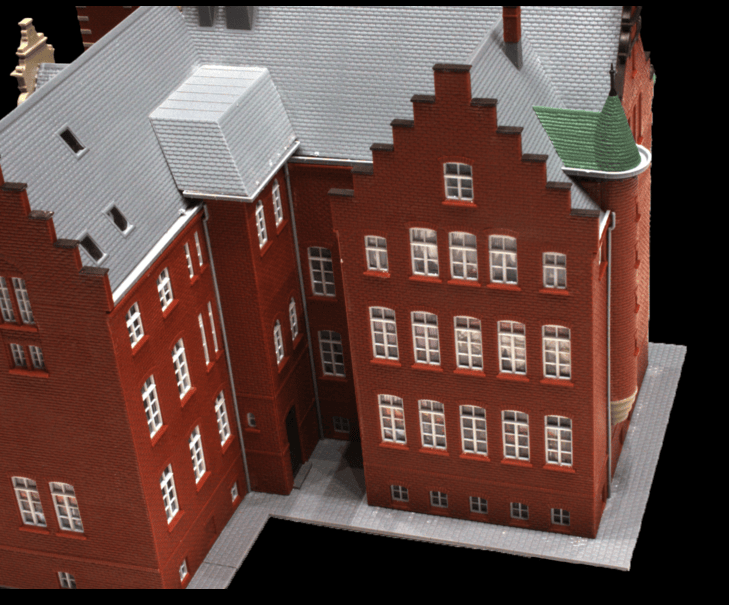}};
        \spy[color=red] on (3,1.8) in node[anchor=south west] at ($(image.south west) + (0.02, 0.03)$);

    \end{tikzpicture} &
        \begin{tikzpicture}[spy using outlines={rectangle, magnification=2, size=1.25cm, connect spies}]
        \node[anchor=south west,inner sep=0] (image) at (0,0) {\includegraphics[width=0.33\linewidth]{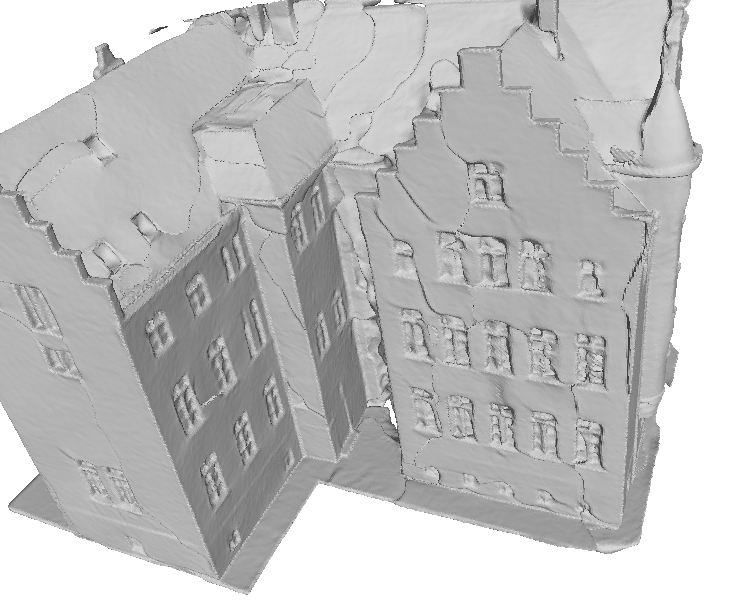}};
        \spy[color=red] on (3.2,1.8) in node[anchor=south west] at ($(image.south west) + (0.02, 0.03)$);

    \end{tikzpicture} &
        \begin{tikzpicture}[spy using outlines={rectangle, magnification=2, size=1.25cm, connect spies}]
        \node[anchor=south west,inner sep=0] (image) at (0,0) {\includegraphics[width=0.33\linewidth]{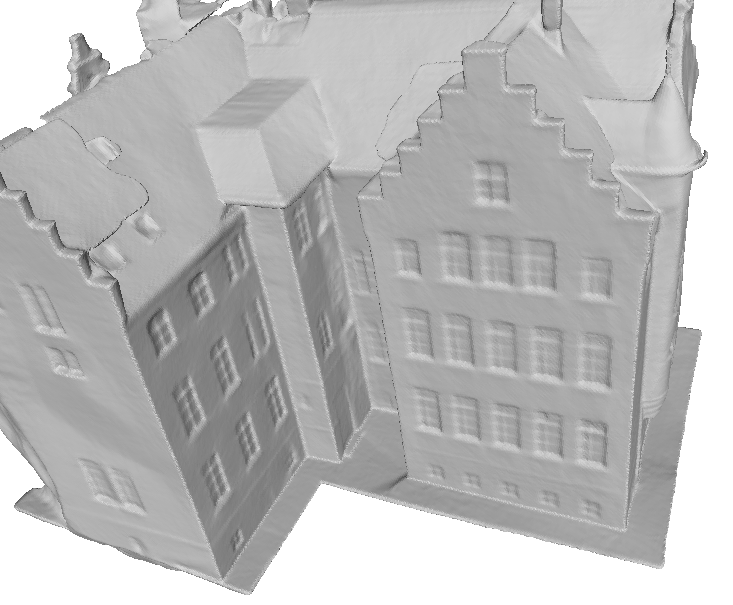}};
        \spy[color=red] on (3.2,1.8) in node[anchor=south west] at ($(image.south west) + (0.02, 0.03)$);

    \end{tikzpicture} \\
        \begin{tikzpicture}[spy using outlines={rectangle, magnification=2, size=1.25cm, connect spies}]
        \node[anchor=south west,inner sep=0] (image) at (0,0) {\includegraphics[width=0.33\linewidth]{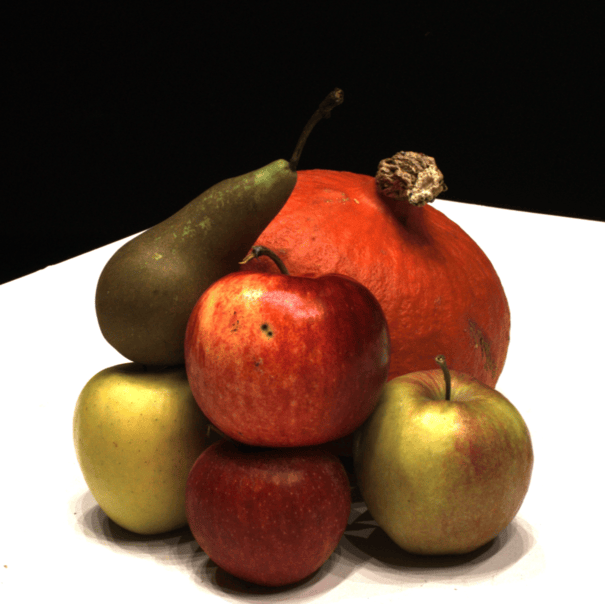}};
        \spy[color=red] on (2.35,2.35) in node[anchor=north west] at ($(image.north west) + (0.02, -0.03)$);
    \end{tikzpicture}
         &        
        \begin{tikzpicture}[spy using outlines={rectangle, magnification=2, size=1.25cm, connect spies}]
        \node[anchor=south west,inner sep=0] (image) at (0,0) {\includegraphics[width=0.33\linewidth]{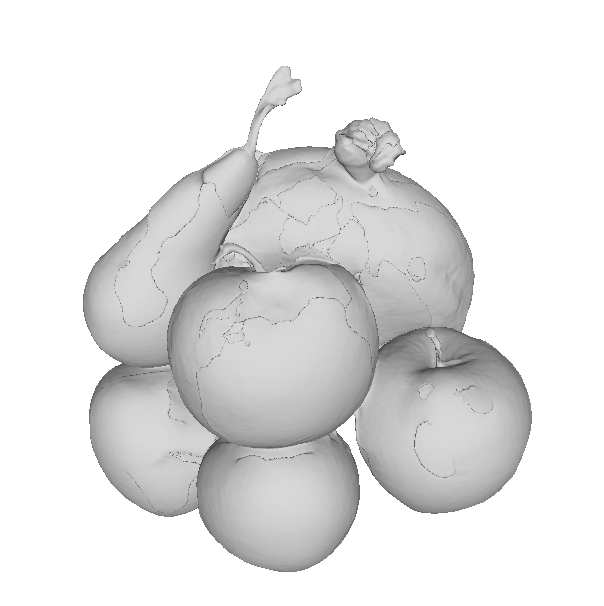}};
        \spy[color=red] on (2.2,2.4) in node[anchor=north west] at ($(image.north west) + (0.02, -0.03) $);
    \end{tikzpicture} &
        \begin{tikzpicture}[spy using outlines={rectangle, magnification=2, size=1.25cm, connect spies}]
        \node[anchor=south west,inner sep=0] (image) at (0,0) {\includegraphics[width=0.33\linewidth]{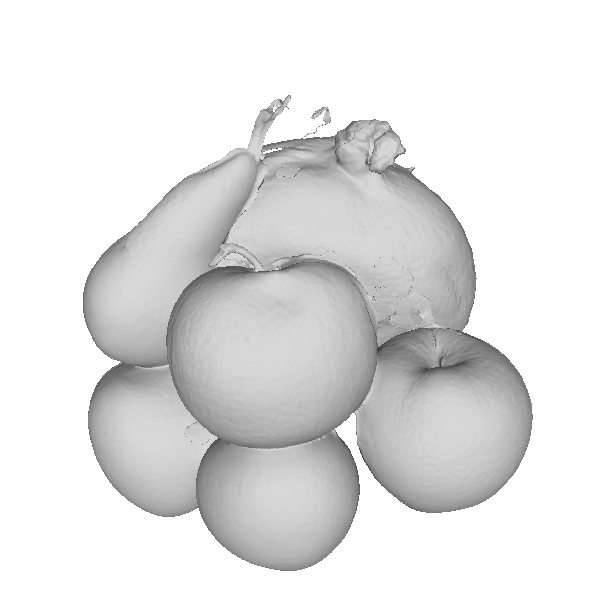}};
        \spy[color=red] on (2.2,2.4) in node[anchor=north west] at ($(image.north west) + (0.02, -0.03)$);

    \end{tikzpicture} \\
        \small{\textbf{Input image}} & \small{\textbf{Single MLP}} & \small{\textbf{Full two-MLP network (ours)}} \\ 
    \end{tabular}}
    
\caption{Ablation study comparing the single‐MLP approach (middle) to our full two-MLP network (right). 
  The single MLP struggles to capture fine details and exhibits noticeable artifacts, 
  whereas our full network yields higher fidelity geometry.}
    \label{fig:ablation_single_network}
\end{figure}

\begin{figure}
    \centering
    \includegraphics[width=0.97\linewidth]{images/reading_diff_views_comp.png}
    \caption{Qualitative comparison of Reading between NeuS~\cite{wang2021neus}, PS-NeRF~\cite{yang2022ps}, RNb-NeuS~\cite{Brument23} and our method for different numbers of input
viewpoints.}
    \label{fig:reading_diff_views}
\end{figure}

\subsection{Ablation study} 
\label{sec:ablation}
We conduct ablation studies to investigate the importance of each component of our method.
\subsubsection{Effect of normal loss}
~\aptLtoX[graphic=no,type=html]{Table~\ref{tab:mergedAblations}a}{\Cref{tab:lossComparison}}  compares the performance of models trained with and without the normal loss component $\mathcal{L}_{\text{dnc}}$. The results indicate that incorporating the normal loss alongside the color loss $\mathcal{L}_{\text{color}}$ significantly reduces the Mean CD from $4.21$ to $0.84$. This demonstrates the importance of the surface normal information for more accurate 3D reconstructions. We do not introduce depth loss because we obtain the relative depth from the Depth Anything~\cite{depthanything}, which may introduce scale ambiguity in depth values.

\subsubsection{Effect of different monocular depth estimators}
\aptLtoX[graphic=no,type=html]{Table~\ref{tab:mergedAblations}b}{\Cref{tab:depthComparison}} analyzes different monocular depth estimators affecting the reconstruction quality. Depth Anything~\cite{depthanything}  achieves a lower Mean CD of $0.84$ compared to $0.97$ for the Omnidata~\cite{eftekhar2021omnidata}. This suggests a slight improvement in the quality of the reconstructed meshes when selecting a more accurate depth estimator.

\subsubsection{Effect of different normal extraction methods}
We also want to analyze the effect of different surface normal extraction methods on the reconstructed meshes. So, we used two different methods to find the rendered normals $\hat{\normalvec}$ from the SDF network, namely (1) the volumetric accumulation method used in NeuRIS~\cite{wang2022neuris} and MonoSDF~\cite{Yu2022MonoSDF} and (2) the proposed surface point localization method described in \Cref{sec:surface_extraction}. 
~\aptLtoX[graphic=no,type=html]{Table~\ref{tab:mergedAblations}c}{\Cref{tab:normalComparison}} shows that our method significantly outperforms the volumetric accumulation along the rays approach.
\subsubsection{Single‐MLP vs.\ full‐network}
\label{sec:ablation_single_network}
To assess the importance of the color MLP, we undertake an ablation study in which we discard the color MLP of our network and retain the first MLP to output both the SDF value and the RGB color. We modify the SDF MLP to take view direction $v$ in addition to 3D point $\mathbf{x}$ as input. Specifically, this single MLP consists of $8$ fully connected layers, each containing $256$ hidden units as before, and uses positional encoding for both the 3D point and the view direction. We train this single MLP using the same loss functions and data settings as our full model.

As shown in \aptLtoX[graphic=no,type=html]{Table~\ref{tab:mergedAblations2}a}{\Cref{tab:single_network}}, the single‐MLP approach yields poor reconstructions on the DTU dataset, with a mean CD of $1.57$, compared to $0.84$ for our full two‐MLP model. Visually, the single‐MLP reconstructions contain artifacts, as shown in \Cref{fig:ablation_single_network}. It is important to note that our original design employs a dedicated neural renderer that maps the latent geometric representation (i.e., the SDF gradient and a 256-dimensional feature vector) to an appearance signal aligned with the input RGB images, ensuring robust gradient flow to each MLP and reach convergence. In the single-MLP variant, this specialized mapping is absent, and the network cannot fully leverage explicit gradient signals, resulting in poorer reconstruction quality.

\subsubsection{Architecture Choices for SDF representation}
We compare our implicit SDF scene representation with an explicit SDF grid scene representation. For explicit SDF grid representation, we use a discretized volume with a resolution of $512 \times 512 \times 512$, where each grid cell stores a vector comprising the SDF value and a feature vector $\hat{z}$. We use trilinear interpolation to query for an arbitrary point $\point$ from the SDF grid. We perform this analysis on a subset of scenes from the DTU dataset (8 scenes) and report the mean CD in~\aptLtoX[graphic=no,type=html]{Table~\ref{tab:mergedAblations2}b}{\Cref{tab:archiChoice}}. The table shows that our implicit SDF MLP achieves a lower mean CD (0.90) than the explicit SDF grid (1.05), indicating improved surface reconstruction accuracy. This performance gain can be attributed to the continuous nature of the SDF MLP, which allows for smoother geometry representation and avoids the discretization artifacts introduced by the grid because of its fixed resolution. These results highlight the advantages of our approach in achieving higher-fidelity reconstructions.

\subsubsection{Optimal SDF network architecture}
We analyze the impact of the network depth by varying the number of layers in the SDF network. The results shown in \Cref{tab:7} indicate that increasing the network's depth improves the reconstruction quality up to a certain point. Specifically, the mean CD decreases from $1.12$ to $0.84$ as linear layers increase from five to eight. However, increasing the number of linear layers beyond $8$ does not consistently improve performance, as seen with 9 and 10 layers, resulting in higher CD values.

\subsubsection{Effect of different number of input views}
Finally, we evaluate how the performance of our method changes with the number of input views ($2, 4, 5, 6, 8, 10, 15$ and $20$ views) and compare it with the existing methods.  
~\Cref{fig:reading_diff_views} show that while the quality of the meshes reconstructed using previous methods deteriorates substantially as input views become sparser, our method outputs high-quality meshes even when the number of input views is as few as two.

\section{Conclusion}
\label{sec:conclusion}
In this paper, we have proposed a new method for multi-view reconstruction of objects even in extremely sparse multi-view stereo scenarios where only as few as two images per object are available. We show that previous state-of-the-art MVS and MVPS methods produce 3D surfaces with missing parts and fail to recover fine details. We have demonstrated by extensive evaluation on synthetic and real-world datasets that incorporating high-order geometry cues in the form of surface normals results in highly accurate 3D surface reconstruction, even in situations where only as few as two RGB (front and back) images are available. 
One significant drawback of our method is its slightly increased computational cost. Thus, in the future, we will explore algorithmic acceleration techniques to improve the computation time.

\begin{acks}
    Aarya Patel is funded by the Prime Minister's Fellowship for Doctoral Research, jointly supported by the Science \& Engineering Research Board (SERB), Department of Science and Technology (DST), Government of India, and Dolby Laboratories India. Hamid Laga is supported in part by the Australian Research Council (ARC) Discovery Grant no. DP220102197.

\end{acks}

\bibliographystyle{ACM-Reference-Format}
\bibliography{sample-base}
\clearpage

\appendix

\section{Effect of the number of views}

\Cref{tab:views} shows, in terms of Chamfer Distance (CD) and normal Mean Average Error (MAE), how our method is affected when reducing the number of views from $15$ to $2$. Compared to the state-of-the-art method on the  DiLiGenT-MV dataset, we can see that our method performs very well when we have a large number of views, outperforming even dense multiview photometric stereo methods (MVS). Notably, unlike other methods, the performance of our method drops only slightly when we reduce the number of views from $15$ to $2$. Note that the views are distributed uniformly around the object in this experiment. Thus, when they are sparse, they have little overlap. Yet, our method can recover the overall 3D shape and surface details with very high accuracy.

 Moreover, it can be visually inferred from~\Cref{fig:chamfer_diff_views} that while the performance of state-of-the-art methods degrades as the input views become sparser, our method shows stable performance even when using as few as two views, with only one RGB image per view. Moreover, the Chamfer distance of our method, when using only two views, significantly outperforms the current best multiview photometric stereo model, namely RNb-NeuS \cite{Brument23}.

\aptLtoX[graphic=no,type=html]{\begin{table*}
\caption{Results of different methods on the  DiLiGenT-MV dataset when using $15, 10, 8, 6, 5$, $4$ and $2$ views. Best results are highlighted as \colorbox{colorFst}{1st}, \colorbox{colorSnd}{2nd} and \colorbox{colorTrd}{3rd}.}
\label{tab:views}
\begin{tabular}{l|*{5}{c}|c|*{5}{c}|c}
    \toprule
    & \multicolumn{6}{c|}{\textbf{Chamfer Dist}($\downarrow$) } &  \multicolumn{6}{c}{\textbf{Normal MAE}($\downarrow$)} \\
    \textbf{Method (with 15 views)}
    & \emph{Bear}  & \emph{Buddha}   & \emph{Cow}   & \emph{Pot2}   & \emph{Reading}  & \emph{Average} 
     & \emph{Bear}  & \emph{Buddha}   & \emph{Cow}   & \emph{Pot2}   & \emph{Reading}  & \emph{Average} 
    \\
    \cline{1-13}
    NeuS~\cite{wang2021neus}
    & 27.94 & 12.63 & 32.99 & 33.58 & 26.36 & 26.70
    & 27.68  & 22.42  & 30.88 & 39.43  & 28.11  & 29.70
    \\
    PS-NeRF~\cite{yang2022ps}
    & \colorbox{colorFst}{7.49} & \colorbox{colorTrd}{10.82} & \colorbox{colorSnd}{15.37} & 8.03 & 20.69 & \colorbox{colorTrd}{12.48}
    & \colorbox{colorTrd}{6.77} & 19.02 & \colorbox{colorTrd}{10.97} & 12.00 & \colorbox{colorTrd}{13.89} & 12.53
    \\
    RNb-NeuS~\cite{Brument23}
    & 36.32 & \colorbox{colorSnd}{8.78} & 39.39 & \colorbox{colorSnd}{5.93} & \colorbox{colorSnd}{14.32} & 20.95
    & \colorbox{colorSnd}{3.45}  & \colorbox{colorSnd}{13.66}  & \colorbox{colorSnd}{4.57} & \colorbox{colorSnd}{6.96}  & \colorbox{colorSnd}{9.31} & \colorbox{colorSnd}{7.59}
    \\
    Ours
    & \colorbox{colorSnd}{9.19} & \colorbox{colorFst}{7.79} & \colorbox{colorFst}{10.15} & \colorbox{colorTrd}{6.35} & \colorbox{colorFst}{8.34} & \colorbox{colorFst}{8.36}
    & \colorbox{colorFst}{1.11}  & \colorbox{colorFst}{4.10}  & \colorbox{colorFst}{1.20}  & \colorbox{colorFst}{2.02}  & \colorbox{colorFst}{4.11}  & \colorbox{colorFst}{2.51} 
    \\
    \cline{1-13}
    & \multicolumn{6}{c|}{\textbf{Chamfer Dist}($\downarrow$) } &  \multicolumn{6}{c}{\textbf{Normal MAE}($\downarrow$)} \\
    \textbf{Method (with 10 views)}
    & \emph{Bear}  & \emph{Buddha}   & \emph{Cow}   & \emph{Pot2}   & \emph{Reading}  & \emph{Average} 
     & \emph{Bear}  & \emph{Buddha}   & \emph{Cow}   & \emph{Pot2}   & \emph{Reading}  & \emph{Average} 
    \\
    \cline{1-13}
    NeuS~\cite{wang2021neus}
    & 38.62 & 15.48 & 55.45 & 47.58 & 25.13 & 27.94
    & 22.12  & 27.88  & 33.18 & 38.34  & 28.16  & 27.94
    \\
    PS-NeRF~\cite{yang2022ps}
    & \colorbox{colorSnd}{9.22} & \colorbox{colorTrd}{10.07} & \colorbox{colorSnd}{13.08} & \colorbox{colorTrd}{8.93} & \colorbox{colorSnd}{12.84} & \colorbox{colorSnd}{10.83}
    & \colorbox{colorTrd}{7.99} & \colorbox{colorTrd}{17.97} & \colorbox{colorTrd}{9.93} & \colorbox{colorTrd}{12.12} & \colorbox{colorTrd}{11.48} & \colorbox{colorTrd}{11.90}
    \\
    RNb-NeuS~\cite{Brument23}
    & 34.40 & \colorbox{colorFst}{6.17} & 46.07 & \colorbox{colorSnd}{8.44} & \colorbox{colorTrd}{17.26} & 22.47
    & \colorbox{colorSnd}{3.41}  & \colorbox{colorSnd}{12.85}  & \colorbox{colorSnd}{4.87} & \colorbox{colorSnd}{6.79}  & \colorbox{colorSnd}{8.95}  & \colorbox{colorSnd}{7.37}
    \\
    Ours
    & \colorbox{colorFst}
{7.89} & \colorbox{colorSnd}{7.71} & \colorbox{colorFst}{12.21} & \colorbox{colorFst}{5.93} & \colorbox{colorFst}{7.87} & \colorbox{colorFst}{8.32}
    & \colorbox{colorFst}{1.20}  & \colorbox{colorFst}{5.00}  & \colorbox{colorFst}{1.33}  & \colorbox{colorFst}{2.08}  & \colorbox{colorFst}{3.97}  & \colorbox{colorFst}{2.72} 
    \\
    \cline{1-13}
    & \multicolumn{6}{c|}{\textbf{Chamfer Dist}($\downarrow$) } &  \multicolumn{6}{c}{\textbf{Normal MAE}($\downarrow$)} \\
    \textbf{Method (with eight views)}
    & \emph{Bear}  & \emph{Buddha}   & \emph{Cow}   & \emph{Pot2}   & \emph{Reading}  & \emph{Average} 
     & \emph{Bear}  & \emph{Buddha}   & \emph{Cow}   & \emph{Pot2}   & \emph{Reading}  & \emph{Average} 
    \\
    \cline{1-13}
    NeuS~\cite{wang2021neus}
    & 36.64 & 18.43 & 61.52 & 46.68 & 41.26 & 40.91
    & 24.71  & 27.59  & 29.39 & 35.68  & 31.84  & 29.84
    \\
    PS-NeRF~\cite{yang2022ps}
    & \colorbox{colorSnd}{8.61} & \colorbox{colorTrd}{10.68} & \colorbox{colorSnd}{15.51} & \colorbox{colorSnd}{8.53} & \colorbox{colorSnd}{15.76} & \colorbox{colorSnd}{11.82}
    & \colorbox{colorTrd}{8.08} & \colorbox{colorTrd}{18.16} & \colorbox{colorTrd}{11.14} & \colorbox{colorTrd}{12.60} & \colorbox{colorTrd}{12.65} & \colorbox{colorTrd}{12.52}
    \\
    RNb-NeuS~\cite{Brument23}
    & 38.31 & \colorbox{colorFst}{7.79} & \colorbox{colorTrd}{46.91} & \colorbox{colorTrd}{12.58} & \colorbox{colorTrd}{17.85} & \colorbox{colorTrd}{24.68}
    & \colorbox{colorSnd}{3.49}  & \colorbox{colorSnd}{12.95}  &\colorbox{colorSnd}{5.05} & \colorbox{colorSnd}{7.79}  & \colorbox{colorSnd}{8.55}  & \colorbox{colorSnd}{7.56}
    \\
    Ours
    & \colorbox{colorFst}{9.09} & \colorbox{colorSnd}{8.06} & \colorbox{colorFst}{12.52} & \colorbox{colorFst}{4.56} & \colorbox{colorFst}{7.35} & \colorbox{colorFst}{8.31}
    & \colorbox{colorFst}{1.08}  & \colorbox{colorFst}{4.22}  & \colorbox{colorFst}{1.27}  & \colorbox{colorFst}{2.09}  & \colorbox{colorFst}{4.67}  & \colorbox{colorFst}{2.66} 
    \\
    \cline{1-13}
    & \multicolumn{6}{c|}{\textbf{Chamfer Dist}($\downarrow$) } &  \multicolumn{6}{c}{\textbf{Normal MAE}($\downarrow$)} \\
    \textbf{Method (with six views)}
    & \emph{Bear}  & \emph{Buddha}   & \emph{Cow}   & \emph{Pot2}   & \emph{Reading}  & \emph{Average} 
     & \emph{Bear}  & \emph{Buddha}   & \emph{Cow}   & \emph{Pot2}   & \emph{Reading}  & \emph{Average} 
    \\
    \cline{1-13}
    NeuS~\cite{wang2021neus}
    & 42.86 & 20.12 & 63.24 & 49.26 & 40.46 & 43.18
    & 21.69  & 30.78  & 27.36 & 32.01  & 31.40  & 28.65
    \\
    PS-NeRF~\cite{yang2022ps}
    & \colorbox{colorSnd}{10.60} & \colorbox{colorTrd}{11.07} & \colorbox{colorSnd}{21.20} & \colorbox{colorSnd}{6.82} & \colorbox{colorTrd}{23.88} & \colorbox{colorSnd}{14.72}
    & \colorbox{colorTrd}{9.35} & \colorbox{colorTrd}{19.73} & \colorbox{colorTrd}{10.87} & \colorbox{colorTrd}{10.06} & \colorbox{colorTrd}{15.04} & \colorbox{colorTrd}{13.01}
    \\
    RNb-NeuS~\cite{Brument23}
    & 34.25 & \colorbox{colorFst}{5.88} & 37.96 & \colorbox{colorTrd}{14.26} & \colorbox{colorSnd}{18.92} & \colorbox{colorTrd}{22.25}
    & \colorbox{colorSnd}{4.60}  & \colorbox{colorSnd}{14.64}  & \colorbox{colorSnd}{8.16} & \colorbox{colorSnd}{7.71}  & \colorbox{colorSnd}{9.87}  & \colorbox{colorSnd}{8.99}
    \\
    Ours
    & \colorbox{colorFst}{8.32} & \colorbox{colorSnd}{8.61} & \colorbox{colorFst}{14.75} & \colorbox{colorFst}{4.10}  & \colorbox{colorFst}{7.64} & \colorbox{colorFst}{8.68}
    & \colorbox{colorFst}{1.16}  & \colorbox{colorFst}{4.47}  & \colorbox{colorFst}{1.36}  & \colorbox{colorFst}{2.22}  & \colorbox{colorFst}{4.78}  & \colorbox{colorFst}{2.79} 
    \\
    \cline{1-13}
    & \multicolumn{6}{c|}{\textbf{Chamfer Dist}($\downarrow$) } &  \multicolumn{6}{c}{\textbf{Normal MAE}($\downarrow$)} \\
    \textbf{Method (5 views)}
    & \emph{Bear}  & \emph{Buddha}   & \emph{Cow}   & \emph{Pot2}   & \emph{Reading}  & \emph{Average} 
     & \emph{Bear}  & \emph{Buddha}   & \emph{Cow}   & \emph{Pot2}   & \emph{Reading}  & \emph{Average} 
    \\
    \cline{1-13}
    NeuS~\cite{wang2021neus}
    & 45.18 & 27.11 & 66.37 & 39.72 & 51.92 & 46.06
    & 22.40  & 36.94  & 33.07 & 33.86  & 33.82  & 32.02
    \\
    PS-NeRF~\cite{yang2022ps}
    & \colorbox{colorSnd}{11.97} & \colorbox{colorTrd}{12.43} & \colorbox{colorSnd}{21.78} & \colorbox{colorTrd}{9.42} & \colorbox{colorSnd}{16.37} & \colorbox{colorSnd}{14.39}
    & \colorbox{colorSnd}{9.21}  & \colorbox{colorTrd}{19.17}  & \colorbox{colorTrd}{12.89} & \colorbox{colorTrd}{13.41}  & \colorbox{colorTrd}{13.97}  & \colorbox{colorSnd}{13.73}
    \\
    RNb-NeuS~\cite{Brument23}
    & \colorbox{colorTrd}{29.00} & \colorbox{colorFst}{7.65} & \colorbox{colorTrd}{41.26} & \colorbox{colorSnd}{9.34} & \colorbox{colorTrd}{23.54} & \colorbox{colorTrd}{22.16}
    & 78.10  & \colorbox{colorSnd}{15.23}  & \colorbox{colorSnd}{6.20} & \colorbox{colorSnd}{8.29}  & \colorbox{colorSnd}{10.92}  & \colorbox{colorTrd}{23.75}
    \\
    Ours
    & \colorbox{colorFst}{8.89} & \colorbox{colorSnd}{8.72} & \colorbox{colorFst}{11.52} & \colorbox{colorFst}{4.56} & \colorbox{colorFst}{7.58} & \colorbox{colorFst}{7.11}
    & \colorbox{colorFst}{1.27}  & \colorbox{colorFst}{4.73}  & \colorbox{colorFst}{1.44}  & \colorbox{colorFst}{2.16}  & \colorbox{colorFst}{4.18}  & \colorbox{colorFst}{2.76} 
    \\
    \cline{1-13}
    & \multicolumn{6}{c|}{\textbf{Chamfer Dist}($\downarrow$) } &  \multicolumn{6}{c}{\textbf{Normal MAE}($\downarrow$)} \\
    \textbf{Method (with four views)}
    & \emph{Bear}  & \emph{Buddha}   & \emph{Cow}   & \emph{Pot2}   & \emph{Reading}  & \emph{Average} 
     & \emph{Bear}  & \emph{Buddha}   & \emph{Cow}   & \emph{Pot2}   & \emph{Reading}  & \emph{Average} 
    \\
    \cline{1-13}
    NeuS~\cite{wang2021neus}
    & 60.76 & 40.10 & 66.14 & \colorbox{colorTrd}{39.52} & 49.18 & 51.14
    & 30.79  & 39.46  & 30.76 & 32.97  & 33.92  & 33.58
    \\
    PS-NeRF~\cite{yang2022ps}
    & \colorbox{colorSnd}{11.09} & \colorbox{colorTrd}{14.87} & \colorbox{colorSnd}{22.21} & 97.99 & \colorbox{colorSnd}{19.34} & \colorbox{colorTrd}{33.10}
    & \colorbox{colorTrd}{9.46} & \colorbox{colorTrd}{23.84} & \colorbox{colorTrd}{12.43} & \colorbox{colorTrd}{11.80} & \colorbox{colorTrd}{14.38} & \colorbox{colorTrd}{14.38}
    \\
    RNb-NeuS~\cite{Brument23}
    & \colorbox{colorTrd}{20.75} & \colorbox{colorSnd}{9.68} & \colorbox{colorTrd}{23.98} & \colorbox{colorSnd}{12.49} & \colorbox{colorTrd}{23.56} & \colorbox{colorSnd}{18.09}
    & \colorbox{colorSnd}{5.45}  & \colorbox{colorSnd}{18.30}  & \colorbox{colorSnd}{6.57} &\colorbox{colorSnd}{8.85}  & \colorbox{colorSnd}{11.16}  & \colorbox{colorSnd}{10.06}
    \\
    Ours
    & \colorbox{colorFst}{8.67} & \colorbox{colorFst}{8.30} & \colorbox{colorFst}{7.78} & \colorbox{colorFst}{4.23} & \colorbox{colorFst}{7.46} & \colorbox{colorFst}{7.28}
    & \colorbox{colorFst}{1.96}  & \colorbox{colorFst}{5.37}  & \colorbox{colorFst}{1.56}  & \colorbox{colorFst}{2.56}  & \colorbox{colorFst}{4.26}  & \colorbox{colorFst}{3.14} 
    \\
    \cline{1-13}
    & \multicolumn{6}{c|}{\textbf{Chamfer Dist}($\downarrow$) } &  \multicolumn{6}{c}{\textbf{Normal MAE}($\downarrow$)} \\
    \textbf{Method (with two views)} 
    & \emph{Bear}  & \emph{Buddha}   & \emph{Cow}   & \emph{Pot2}   & \emph{Reading}  & \emph{Average} 
     & \emph{Bear}  & \emph{Buddha}   & \emph{Cow}   & \emph{Pot2}   & \emph{Reading}  & \emph{Average} 
    \\
    \cline{1-13}
    NeuS \cite{wang2021neus}
    & 142.44 & 71.55 & 130.19 & 131.58 & 156.26 & 126.40
    & 35.84  & \colorbox{colorSnd}{37.23}  & 32.25 & 32.37  & 42.85  & 36.11
    \\
    PS-NeRF~\cite{yang2022ps}
    & \colorbox{colorSnd}{29.26} & \colorbox{colorTrd}{50.42} & \colorbox{colorSnd}{37.21} & \colorbox{colorSnd}{18.72} & \colorbox{colorTrd}{56.81} & \colorbox{colorSnd}{38.48}
    & \colorbox{colorSnd}{15.59}  & \colorbox{colorTrd}{43.16}  & \colorbox{colorTrd}{17.35} & \colorbox{colorSnd}{14.94}  & \colorbox{colorTrd}{31.40}  & \colorbox{colorSnd}{24.45}
    \\
    RNb-NeuS~\cite{Brument23}
    & \colorbox{colorTrd}{39.08} & \colorbox{colorSnd}{45.58} & \colorbox{colorTrd}{48.87} & \colorbox{colorTrd}{23.18} & \colorbox{colorFst}{43.38} & \colorbox{colorTrd}{40.01}
    & \colorbox{colorTrd}{22.95}  & 46.19  & \colorbox{colorSnd}{15.36} & \colorbox{colorTrd}{15.83}  & \colorbox{colorSnd}{22.52}  & \colorbox{colorTrd}{24.57}
    \\
    Ours
    & \colorbox{colorFst}{17.49} & \colorbox{colorFst}{11.05} & \colorbox{colorFst}{11.40} & \colorbox{colorFst}{7.32} & \colorbox{colorSnd}{61.69} & \colorbox{colorFst}{21.79}
    & \colorbox{colorFst}{2.54}  & \colorbox{colorFst}{6.25}  & \colorbox{colorFst}{1.69}  & \colorbox{colorFst}{3.11}  & \colorbox{colorFst}{5.67}  & \colorbox{colorFst}{3.85} 
    \\
    \bottomrule
\end{tabular}
\end{table*}}{\begin{table*}[t]
\caption{Results of different methods on the  DiLiGenT-MV dataset when using $15, 10, 8, 6, 5$, $4$ and $2$ views. Best results are highlighted as \colorbox{colorFst}{1st}, \colorbox{colorSnd}{2nd} and \colorbox{colorTrd}{3rd}.}
\label{tab:views}
\resizebox{\textwidth}{!}{
\begin{tabular}{l|*{5}{c}|c|*{5}{c}|c}
    \toprule
    & \multicolumn{6}{c|}{\textbf{Chamfer Dist}($\downarrow$) } &  \multicolumn{6}{c}{\textbf{Normal MAE}($\downarrow$)} \\
    \textbf{Method (with 15 views)}
    & \emph{Bear}  & \emph{Buddha}   & \emph{Cow}   & \emph{Pot2}   & \emph{Reading}  & \emph{Average} 
     & \emph{Bear}  & \emph{Buddha}   & \emph{Cow}   & \emph{Pot2}   & \emph{Reading}  & \emph{Average} 
    \\
    \hline
    NeuS~\cite{wang2021neus}
    & 27.94 & 12.63 & 32.99 & 33.58 & 26.36 & 26.70
    & 27.68  & 22.42  & 30.88 & 39.43  & 28.11  & 29.70
    \\
    PS-NeRF~\cite{yang2022ps}
    & \colorbox{colorFst}{7.49} & \colorbox{colorTrd}{10.82} & \colorbox{colorSnd}{15.37} & 8.03 & 20.69 & \colorbox{colorTrd}{12.48}
    & \colorbox{colorTrd}{6.77} & 19.02 & \colorbox{colorTrd}{10.97} & 12.00 & \colorbox{colorTrd}{13.89} & 12.53
    \\
    RNb-NeuS~\cite{Brument23}
    & 36.32 & \colorbox{colorSnd}{8.78} & 39.39 & \colorbox{colorSnd}{5.93} & \colorbox{colorSnd}{14.32} & 20.95
    & \colorbox{colorSnd}{3.45}  & \colorbox{colorSnd}{13.66}  & \colorbox{colorSnd}{4.57} & \colorbox{colorSnd}{6.96}  & \colorbox{colorSnd}{9.31} & \colorbox{colorSnd}{7.59}
    \\
    Ours
    & \colorbox{colorSnd}{9.19} & \colorbox{colorFst}{7.79} & \colorbox{colorFst}{10.15} & \colorbox{colorTrd}{6.35} & \colorbox{colorFst}{8.34} & \colorbox{colorFst}{8.36}
    & \colorbox{colorFst}{1.11}  & \colorbox{colorFst}{4.10}  & \colorbox{colorFst}{1.20}  & \colorbox{colorFst}{2.02}  & \colorbox{colorFst}{4.11}  & \colorbox{colorFst}{2.51} 
    \\
    \bottomrule
    \multicolumn{13}{c}{}\\
\end{tabular}}
\resizebox{\textwidth}{!}{
\begin{tabular}{l|*{5}{c}|c|*{5}{c}|c}
    \toprule
    & \multicolumn{6}{c|}{\textbf{Chamfer Dist}($\downarrow$) } &  \multicolumn{6}{c}{\textbf{Normal MAE}($\downarrow$)} \\
    \textbf{Method (with 10 views)}
    & \emph{Bear}  & \emph{Buddha}   & \emph{Cow}   & \emph{Pot2}   & \emph{Reading}  & \emph{Average} 
     & \emph{Bear}  & \emph{Buddha}   & \emph{Cow}   & \emph{Pot2}   & \emph{Reading}  & \emph{Average} 
    \\
    \hline
    NeuS~\cite{wang2021neus}
    & 38.62 & 15.48 & 55.45 & 47.58 & 25.13 & 27.94
    & 22.12  & 27.88  & 33.18 & 38.34  & 28.16  & 27.94
    \\
    PS-NeRF~\cite{yang2022ps}
    & \colorbox{colorSnd}{9.22} & \colorbox{colorTrd}{10.07} & \colorbox{colorSnd}{13.08} & \colorbox{colorTrd}{8.93} & \colorbox{colorSnd}{12.84} & \colorbox{colorSnd}{10.83}
    & \colorbox{colorTrd}{7.99} & \colorbox{colorTrd}{17.97} & \colorbox{colorTrd}{9.93} & \colorbox{colorTrd}{12.12} & \colorbox{colorTrd}{11.48} & \colorbox{colorTrd}{11.90}
    \\
    RNb-NeuS~\cite{Brument23}
    & 34.40 & \colorbox{colorFst}{6.17} & 46.07 & \colorbox{colorSnd}{8.44} & \colorbox{colorTrd}{17.26} & 22.47
    & \colorbox{colorSnd}{3.41}  & \colorbox{colorSnd}{12.85}  & \colorbox{colorSnd}{4.87} & \colorbox{colorSnd}{6.79}  & \colorbox{colorSnd}{8.95}  & \colorbox{colorSnd}{7.37}
    \\
    Ours
    & \colorbox{colorFst}
{7.89} & \colorbox{colorSnd}{7.71} & \colorbox{colorFst}{12.21} & \colorbox{colorFst}{5.93} & \colorbox{colorFst}{7.87} & \colorbox{colorFst}{8.32}
    & \colorbox{colorFst}{1.20}  & \colorbox{colorFst}{5.00}  & \colorbox{colorFst}{1.33}  & \colorbox{colorFst}{2.08}  & \colorbox{colorFst}{3.97}  & \colorbox{colorFst}{2.72} 
    \\
    \bottomrule
    \multicolumn{13}{c}{}\\
\end{tabular}}
\resizebox{\textwidth}{!}{
\begin{tabular}{l|*{5}{c}|c|*{5}{c}|c}
    \toprule
    & \multicolumn{6}{c|}{\textbf{Chamfer Dist}($\downarrow$) } &  \multicolumn{6}{c}{\textbf{Normal MAE}($\downarrow$)} \\
    \textbf{Method (with eight views)}
    & \emph{Bear}  & \emph{Buddha}   & \emph{Cow}   & \emph{Pot2}   & \emph{Reading}  & \emph{Average} 
     & \emph{Bear}  & \emph{Buddha}   & \emph{Cow}   & \emph{Pot2}   & \emph{Reading}  & \emph{Average} 
    \\
    \hline
    NeuS~\cite{wang2021neus}
    & 36.64 & 18.43 & 61.52 & 46.68 & 41.26 & 40.91
    & 24.71  & 27.59  & 29.39 & 35.68  & 31.84  & 29.84
    \\
    PS-NeRF~\cite{yang2022ps}
    & \colorbox{colorSnd}{8.61} & \colorbox{colorTrd}{10.68} & \colorbox{colorSnd}{15.51} & \colorbox{colorSnd}{8.53} & \colorbox{colorSnd}{15.76} & \colorbox{colorSnd}{11.82}
    & \colorbox{colorTrd}{8.08} & \colorbox{colorTrd}{18.16} & \colorbox{colorTrd}{11.14} & \colorbox{colorTrd}{12.60} & \colorbox{colorTrd}{12.65} & \colorbox{colorTrd}{12.52}
    \\
    RNb-NeuS~\cite{Brument23}
    & 38.31 & \colorbox{colorFst}{7.79} & \colorbox{colorTrd}{46.91} & \colorbox{colorTrd}{12.58} & \colorbox{colorTrd}{17.85} & \colorbox{colorTrd}{24.68}
    & \colorbox{colorSnd}{3.49}  & \colorbox{colorSnd}{12.95}  &\colorbox{colorSnd}{5.05} & \colorbox{colorSnd}{7.79}  & \colorbox{colorSnd}{8.55}  & \colorbox{colorSnd}{7.56}
    \\
    Ours
    & \colorbox{colorFst}{9.09} & \colorbox{colorSnd}{8.06} & \colorbox{colorFst}{12.52} & \colorbox{colorFst}{4.56} & \colorbox{colorFst}{7.35} & \colorbox{colorFst}{8.31}
    & \colorbox{colorFst}{1.08}  & \colorbox{colorFst}{4.22}  & \colorbox{colorFst}{1.27}  & \colorbox{colorFst}{2.09}  & \colorbox{colorFst}{4.67}  & \colorbox{colorFst}{2.66} 
    \\
    \bottomrule
    \multicolumn{13}{c}{}\\
\end{tabular}}
\resizebox{\textwidth}{!}{
\begin{tabular}{l|*{5}{c}|c|*{5}{c}|c}
    \toprule
    & \multicolumn{6}{c|}{\textbf{Chamfer Dist}($\downarrow$) } &  \multicolumn{6}{c}{\textbf{Normal MAE}($\downarrow$)} \\
    \textbf{Method (with six views)}
    & \emph{Bear}  & \emph{Buddha}   & \emph{Cow}   & \emph{Pot2}   & \emph{Reading}  & \emph{Average} 
     & \emph{Bear}  & \emph{Buddha}   & \emph{Cow}   & \emph{Pot2}   & \emph{Reading}  & \emph{Average} 
    \\
    \hline
    NeuS~\cite{wang2021neus}
    & 42.86 & 20.12 & 63.24 & 49.26 & 40.46 & 43.18
    & 21.69  & 30.78  & 27.36 & 32.01  & 31.40  & 28.65
    \\
    PS-NeRF~\cite{yang2022ps}
    & \colorbox{colorSnd}{10.60} & \colorbox{colorTrd}{11.07} & \colorbox{colorSnd}{21.20} & \colorbox{colorSnd}{6.82} & \colorbox{colorTrd}{23.88} & \colorbox{colorSnd}{14.72}
    & \colorbox{colorTrd}{9.35} & \colorbox{colorTrd}{19.73} & \colorbox{colorTrd}{10.87} & \colorbox{colorTrd}{10.06} & \colorbox{colorTrd}{15.04} & \colorbox{colorTrd}{13.01}
    \\
    RNb-NeuS~\cite{Brument23}
    & 34.25 & \colorbox{colorFst}{5.88} & 37.96 & \colorbox{colorTrd}{14.26} & \colorbox{colorSnd}{18.92} & \colorbox{colorTrd}{22.25}
    & \colorbox{colorSnd}{4.60}  & \colorbox{colorSnd}{14.64}  & \colorbox{colorSnd}{8.16} & \colorbox{colorSnd}{7.71}  & \colorbox{colorSnd}{9.87}  & \colorbox{colorSnd}{8.99}
    \\
    Ours
    & \colorbox{colorFst}{8.32} & \colorbox{colorSnd}{8.61} & \colorbox{colorFst}{14.75} & \colorbox{colorFst}{4.10}  & \colorbox{colorFst}{7.64} & \colorbox{colorFst}{8.68}
    & \colorbox{colorFst}{1.16}  & \colorbox{colorFst}{4.47}  & \colorbox{colorFst}{1.36}  & \colorbox{colorFst}{2.22}  & \colorbox{colorFst}{4.78}  & \colorbox{colorFst}{2.79} 
    \\
    \bottomrule
    \multicolumn{13}{c}{}\\
\end{tabular}}
\resizebox{\textwidth}{!}{
\begin{tabular}{l|*{5}{c}|c|*{5}{c}|c}
    \toprule
    & \multicolumn{6}{c|}{\textbf{Chamfer Dist}($\downarrow$) } &  \multicolumn{6}{c}{\textbf{Normal MAE}($\downarrow$)} \\
    \textbf{Method (5 views)}
    & \emph{Bear}  & \emph{Buddha}   & \emph{Cow}   & \emph{Pot2}   & \emph{Reading}  & \emph{Average} 
     & \emph{Bear}  & \emph{Buddha}   & \emph{Cow}   & \emph{Pot2}   & \emph{Reading}  & \emph{Average} 
    \\
    \hline
    NeuS~\cite{wang2021neus}
    & 45.18 & 27.11 & 66.37 & 39.72 & 51.92 & 46.06
    & 22.40  & 36.94  & 33.07 & 33.86  & 33.82  & 32.02
    \\
    PS-NeRF~\cite{yang2022ps}
    & \colorbox{colorSnd}{11.97} & \colorbox{colorTrd}{12.43} & \colorbox{colorSnd}{21.78} & \colorbox{colorTrd}{9.42} & \colorbox{colorSnd}{16.37} & \colorbox{colorSnd}{14.39}
    & \colorbox{colorSnd}{9.21}  & \colorbox{colorTrd}{19.17}  & \colorbox{colorTrd}{12.89} & \colorbox{colorTrd}{13.41}  & \colorbox{colorTrd}{13.97}  & \colorbox{colorSnd}{13.73}
    \\
    RNb-NeuS~\cite{Brument23}
    & \colorbox{colorTrd}{29.00} & \colorbox{colorFst}{7.65} & \colorbox{colorTrd}{41.26} & \colorbox{colorSnd}{9.34} & \colorbox{colorTrd}{23.54} & \colorbox{colorTrd}{22.16}
    & 78.10  & \colorbox{colorSnd}{15.23}  & \colorbox{colorSnd}{6.20} & \colorbox{colorSnd}{8.29}  & \colorbox{colorSnd}{10.92}  & \colorbox{colorTrd}{23.75}
    \\
    Ours
    & \colorbox{colorFst}{8.89} & \colorbox{colorSnd}{8.72} & \colorbox{colorFst}{11.52} & \colorbox{colorFst}{4.56} & \colorbox{colorFst}{7.58} & \colorbox{colorFst}{7.11}
    & \colorbox{colorFst}{1.27}  & \colorbox{colorFst}{4.73}  & \colorbox{colorFst}{1.44}  & \colorbox{colorFst}{2.16}  & \colorbox{colorFst}{4.18}  & \colorbox{colorFst}{2.76} 
    \\
    \bottomrule
    \multicolumn{13}{c}{}\\
\end{tabular}}
\resizebox{\textwidth}{!}{
\begin{tabular}{l|*{5}{c}|c|*{5}{c}|c}
    \toprule
    & \multicolumn{6}{c|}{\textbf{Chamfer Dist}($\downarrow$) } &  \multicolumn{6}{c}{\textbf{Normal MAE}($\downarrow$)} \\
    \textbf{Method (with four views)}
    & \emph{Bear}  & \emph{Buddha}   & \emph{Cow}   & \emph{Pot2}   & \emph{Reading}  & \emph{Average} 
     & \emph{Bear}  & \emph{Buddha}   & \emph{Cow}   & \emph{Pot2}   & \emph{Reading}  & \emph{Average} 
    \\
    \hline
    NeuS~\cite{wang2021neus}
    & 60.76 & 40.10 & 66.14 & \colorbox{colorTrd}{39.52} & 49.18 & 51.14
    & 30.79  & 39.46  & 30.76 & 32.97  & 33.92  & 33.58
    \\
    PS-NeRF~\cite{yang2022ps}
    & \colorbox{colorSnd}{11.09} & \colorbox{colorTrd}{14.87} & \colorbox{colorSnd}{22.21} & 97.99 & \colorbox{colorSnd}{19.34} & \colorbox{colorTrd}{33.10}
    & \colorbox{colorTrd}{9.46} & \colorbox{colorTrd}{23.84} & \colorbox{colorTrd}{12.43} & \colorbox{colorTrd}{11.80} & \colorbox{colorTrd}{14.38} & \colorbox{colorTrd}{14.38}
    \\
    RNb-NeuS~\cite{Brument23}
    & \colorbox{colorTrd}{20.75} & \colorbox{colorSnd}{9.68} & \colorbox{colorTrd}{23.98} & \colorbox{colorSnd}{12.49} & \colorbox{colorTrd}{23.56} & \colorbox{colorSnd}{18.09}
    & \colorbox{colorSnd}{5.45}  & \colorbox{colorSnd}{18.30}  & \colorbox{colorSnd}{6.57} &\colorbox{colorSnd}{8.85}  & \colorbox{colorSnd}{11.16}  & \colorbox{colorSnd}{10.06}
    \\
    Ours
    & \colorbox{colorFst}{8.67} & \colorbox{colorFst}{8.30} & \colorbox{colorFst}{7.78} & \colorbox{colorFst}{4.23} & \colorbox{colorFst}{7.46} & \colorbox{colorFst}{7.28}
    & \colorbox{colorFst}{1.96}  & \colorbox{colorFst}{5.37}  & \colorbox{colorFst}{1.56}  & \colorbox{colorFst}{2.56}  & \colorbox{colorFst}{4.26}  & \colorbox{colorFst}{3.14} 
    \\
    \bottomrule
\end{tabular}}
\resizebox{\textwidth}{!}{
\begin{tabular}{@{}l|*{5}{c}|c|*{5}{c}|c@{}}
    \toprule
    & \multicolumn{6}{c|}{\textbf{Chamfer Dist}($\downarrow$) } &  \multicolumn{6}{c}{\textbf{Normal MAE}($\downarrow$)} \\
    \textbf{Method (with two views)} 
    & \emph{Bear}  & \emph{Buddha}   & \emph{Cow}   & \emph{Pot2}   & \emph{Reading}  & \emph{Average} 
     & \emph{Bear}  & \emph{Buddha}   & \emph{Cow}   & \emph{Pot2}   & \emph{Reading}  & \emph{Average} 
    \\
    \hline
    NeuS \cite{wang2021neus}
    & 142.44 & 71.55 & 130.19 & 131.58 & 156.26 & 126.40
    & 35.84  & \colorbox{colorSnd}{37.23}  & 32.25 & 32.37  & 42.85  & 36.11
    \\
    PS-NeRF~\cite{yang2022ps}
    & \colorbox{colorSnd}{29.26} & \colorbox{colorTrd}{50.42} & \colorbox{colorSnd}{37.21} & \colorbox{colorSnd}{18.72} & \colorbox{colorTrd}{56.81} & \colorbox{colorSnd}{38.48}
    & \colorbox{colorSnd}{15.59}  & \colorbox{colorTrd}{43.16}  & \colorbox{colorTrd}{17.35} & \colorbox{colorSnd}{14.94}  & \colorbox{colorTrd}{31.40}  & \colorbox{colorSnd}{24.45}
    \\
    RNb-NeuS~\cite{Brument23}
    & \colorbox{colorTrd}{39.08} & \colorbox{colorSnd}{45.58} & \colorbox{colorTrd}{48.87} & \colorbox{colorTrd}{23.18} & \colorbox{colorFst}{43.38} & \colorbox{colorTrd}{40.01}
    & \colorbox{colorTrd}{22.95}  & 46.19  & \colorbox{colorSnd}{15.36} & \colorbox{colorTrd}{15.83}  & \colorbox{colorSnd}{22.52}  & \colorbox{colorTrd}{24.57}
    \\
    Ours
    & \colorbox{colorFst}{17.49} & \colorbox{colorFst}{11.05} & \colorbox{colorFst}{11.40} & \colorbox{colorFst}{7.32} & \colorbox{colorSnd}{61.69} & \colorbox{colorFst}{21.79}
    & \colorbox{colorFst}{2.54}  & \colorbox{colorFst}{6.25}  & \colorbox{colorFst}{1.69}  & \colorbox{colorFst}{3.11}  & \colorbox{colorFst}{5.67}  & \colorbox{colorFst}{3.85} 
    \\
    \bottomrule
\end{tabular}}
\end{table*}}

\section{Additional results}


\begin{figure}[!ht]
    \centering
	\begin{tabular}{cc}
	\begin{tikzpicture}[scale=.7, every mark/.append style={mark size=1pt}] 
	\begin{axis}[width = 8cm, height=6cm, axis y line = left, axis x line = bottom,
        enlarge x limits=true, enlarge y limits=true, axis line style={-Latex[round]},
        legend style={at={(1,1)}, anchor=north east, fill=gray!30!white, draw=none}, legend columns=1,
        xlabel=No. of views, ylabel=Chamfer distance,
        xmajorgrids=true, ymajorgrids=true, yminorgrids=true,
        xticklabels = {2, 4, 5, 6, 8, 10, 15, 20}, xtick={2, 4, 5, 6, 8, 10, 15, 20}]
    \addplot[thick, color=blue,mark=*] plot coordinates {(2, 142.44) (4, 60.76) (5, 45.18) (6, 42.86) (8, 36.64) (10, 38.62) (15, 27.94) (20, 35.02)}; \addlegendentry{NeuS}
    \addplot[thick, color=red,mark=*] plot coordinates {(2, 29.26) (4, 11.09) (5, 11.97) (6, 10.60) (8, 8.61) (10, 9.22) (15, 7.49) (20, 8.65)}; \addlegendentry{PS-NeRF}
    \addplot[thick, color=green,mark=*] plot coordinates {(2, 39.08) (4, 20.75) (5, 29.00) (6, 34.25) (8, 38.31) (10, 34.40) (15, 36.32) (20, 38.19)}; \addlegendentry{RNb-NeuS}    
    \addplot[very thick, color=black,mark=*] plot coordinates {(2, 17.49) (4, 8.67) (5, 8.89) (6, 8.32) (8, 8.09) (10, 7.89) (15, 7.42) (20, 7.07)}; \addlegendentry{Ours}
	\end{axis}
	\end{tikzpicture} &
	\begin{tikzpicture}[scale=.7, every mark/.append style={mark size=1pt}] 
	\begin{axis}[width = 8cm, height=6cm, axis y line = left, axis x line = bottom,
        enlarge x limits=true, enlarge y limits=true, axis line style={-Latex[round]},
        legend style={at={(1,1)}, anchor=north east, fill=gray!30!white, draw=none}, legend columns=1,
        xlabel=No. of views, ylabel=Chamfer distance,
        xmajorgrids=true, ymajorgrids=true, yminorgrids=true,        
        xticklabels = {2, 4, 5, 6, 8, 10, 15, 20}, xtick={2, 4, 5, 6, 8, 10, 15, 20}]
    \addplot[thick, color=blue,mark=*] plot coordinates {(2, 71.55) (4, 40.10) (5, 27.11) (6, 20.12) (8, 18.43) (10, 15.48) (15, 12.63) (20, 10.64)}; \addlegendentry{NeuS}
    \addplot[thick, color=red,mark=*] plot coordinates {(2, 50.42) (4, 14.87) (5, 12.43) (6, 11.07) (8, 10.68) (10, 10.07) (15, 10.82) (20, 8.61)}; \addlegendentry{PS-NeRF}
    \addplot[thick, color=green,mark=*] plot coordinates {(2, 45.58) (4, 9.68) (5, 7.65) (6, 5.88) (8, 7.79) (10, 6.17) (15, 8.78) (20, 7.69)}; \addlegendentry{RNb-NeuS}    
    \addplot[very thick, color=black,mark=*] plot coordinates {(2, 11.05) (4, 8.30) (5, 8.72) (6, 8.61) (8, 8.06) (10, 7.71) (15, 7.79) (20, 4.82)}; \addlegendentry{Ours}
	\end{axis}
	\end{tikzpicture} \\
    \footnotesize (a) Bear & \footnotesize (b) Buddha \\
	\begin{tikzpicture}[scale=.7, every mark/.append style={mark size=1pt}] 
	\begin{axis}[width = 8cm, height=6cm, axis y line = left, axis x line = bottom,
        enlarge x limits=true, enlarge y limits=true, axis line style={-Latex[round]},
        legend style={at={(1,1)}, anchor=north east, fill=gray!30!white, draw=none}, legend columns=1,
        xlabel=No. of views, ylabel=Chamfer distance,
        xmajorgrids=true, ymajorgrids=true, yminorgrids=true,        
        xticklabels = {2, 4, 5, 6, 8, 10, 15, 20}, xtick={2, 4, 5, 6, 8, 10, 15, 20}]
    \addplot[thick, color=blue,mark=*] plot coordinates {(2, 130.19) (4, 66.14) (5, 66.37) (6, 63.24) (8, 61.52) (10, 55.45) (15, 32.99) (20, 27.07)}; \addlegendentry{NeuS}
    \addplot[thick, color=red,mark=*] plot coordinates {(2, 37.21) (4, 22.21) (5, 21.78) (6, 21.20) (8, 15.51) (10, 13.08) (15, 15.37) (20, 10.21)}; \addlegendentry{PS-NeRF}
    \addplot[thick, color=green,mark=*] plot coordinates {(2, 48.87) (4, 23.98) (5, 41.26) (6, 37.96) (8, 46.91) (10, 46.07) (15, 39.39) (20, 42.78)}; \addlegendentry{RNb-NeuS}    
    \addplot[very thick, color=black,mark=*] plot coordinates {(2, 11.40) (4, 7.78) (5, 11.52) (6, 13.46) (8, 12.30) (10, 12.21) (15, 10.51) (20, 8.99)}; \addlegendentry{Ours}
	\end{axis}
	\end{tikzpicture} &
	\begin{tikzpicture}[scale=.7, every mark/.append style={mark size=1pt}] 
	\begin{axis}[width = 8cm, height=6cm, axis y line = left, axis x line = bottom,
        enlarge x limits=true, enlarge y limits=true, axis line style={-Latex[round]},
        legend style={at={(1,1)}, anchor=north east, fill=gray!30!white, draw=none}, legend columns=1,
        xlabel=No. of views, ylabel=Chamfer distance,
        xmajorgrids=true, ymajorgrids=true, yminorgrids=true,        
        xticklabels = {2, 4, 5, 6, 8, 10, 15, 20}, xtick={2, 4, 5, 6, 8, 10, 15, 20}]
    \addplot[thick, color=blue,mark=*] plot coordinates {(2, 131.58) (4, 39.52) (5, 39.72) (6, 49.26) (8, 46.68) (10, 47.58) (15, 33.58) (20, 34.59)}; \addlegendentry{NeuS}
    \addplot[thick, color=red,mark=*] plot coordinates {(2, 18.72) (4, 97.99) (5, 9.42) (6, 6.82) (8, 8.53) (10, 8.93) (15, 8.03) (20, 6.11)}; \addlegendentry{PS-NeRF}
    \addplot[thick, color=green,mark=*] plot coordinates {(2, 23.18) (4, 12.49) (5, 9.34) (6, 14.26) (8, 12.58) (10, 8.44) (15, 5.93) (20, 7.68)}; \addlegendentry{RNb-NeuS}    
    \addplot[very thick, color=black,mark=*] plot coordinates {(2, 5.32) (4, 4.11) (5, 4.56) (6, 4.10) (8, 4.56) (10, 5.93) (15, 6.35) (20, 5.54)}; \addlegendentry{Ours}
	\end{axis}
	\end{tikzpicture} \\
        \footnotesize (c) Cow & \footnotesize (d) Pot2 \\
	\begin{tikzpicture}[scale=.7, every mark/.append style={mark size=1pt}] 
	\begin{axis}[width = 8cm, height=6cm, axis y line = left, axis x line = bottom,
        enlarge x limits=true, enlarge y limits=true, axis line style={-Latex[round]},
        legend style={at={(1,1)}, anchor=north east, fill=gray!30!white, draw=none}, legend columns=1,
        xlabel=No. of views, ylabel=Chamfer distance,
        xmajorgrids=true, ymajorgrids=true, yminorgrids=true,        
        xticklabels = {2, 4, 5, 6, 8, 10, 15, 20}, xtick={2, 4, 5, 6, 8, 10, 15, 20}]
    \addplot[thick, color=blue,mark=*] plot coordinates {(2, 156.26) (4, 49.18) (5, 51.92) (6, 40.46) (8, 41.26) (10, 25.13) (15, 26.36) (20, 14.88)}; \addlegendentry{NeuS}
    \addplot[thick, color=red,mark=*] plot coordinates {(2, 56.81) (4, 19.34) (5, 16.37) (6, 23.88) (8, 15.76) (10, 12.84) (15, 20.69) (20, 12.35)}; \addlegendentry{PS-NeRF}
    \addplot[thick, color=green,mark=*] plot coordinates {(2, 43.38) (4, 23.56) (5, 23.54) (6, 18.92) (8, 17.85) (10, 17.26) (15, 14.32) (20, 15.57)}; \addlegendentry{RNb-NeuS}    
    \addplot[very thick, color=black,mark=*] plot coordinates {(2, 61.69) (4, 7.46) (5, 7.58) (6, 7.64) (8, 7.35) (10, 7.87) (15, 8.34) (20, 7.60)}; \addlegendentry{Ours}
	\end{axis}
	\end{tikzpicture} & \\
	\footnotesize (e) Reading & 
	\end{tabular}    
     \caption{Quantitative comparison, on the DiLiGenT-MV dataset, of the performance of different methods when varying the number of input viewpoints from $2$ to $20$. Note that MVS methods such as ours and NeuS use one RGB image per view while MVPS methods use multiple images per view, each image is captured under different lighting conditions.}
    \label{fig:chamfer_diff_views}
\end{figure}

\begin{figure}
    \centering
    \scalebox{0.9}{\includegraphics[width=0.8\textwidth]{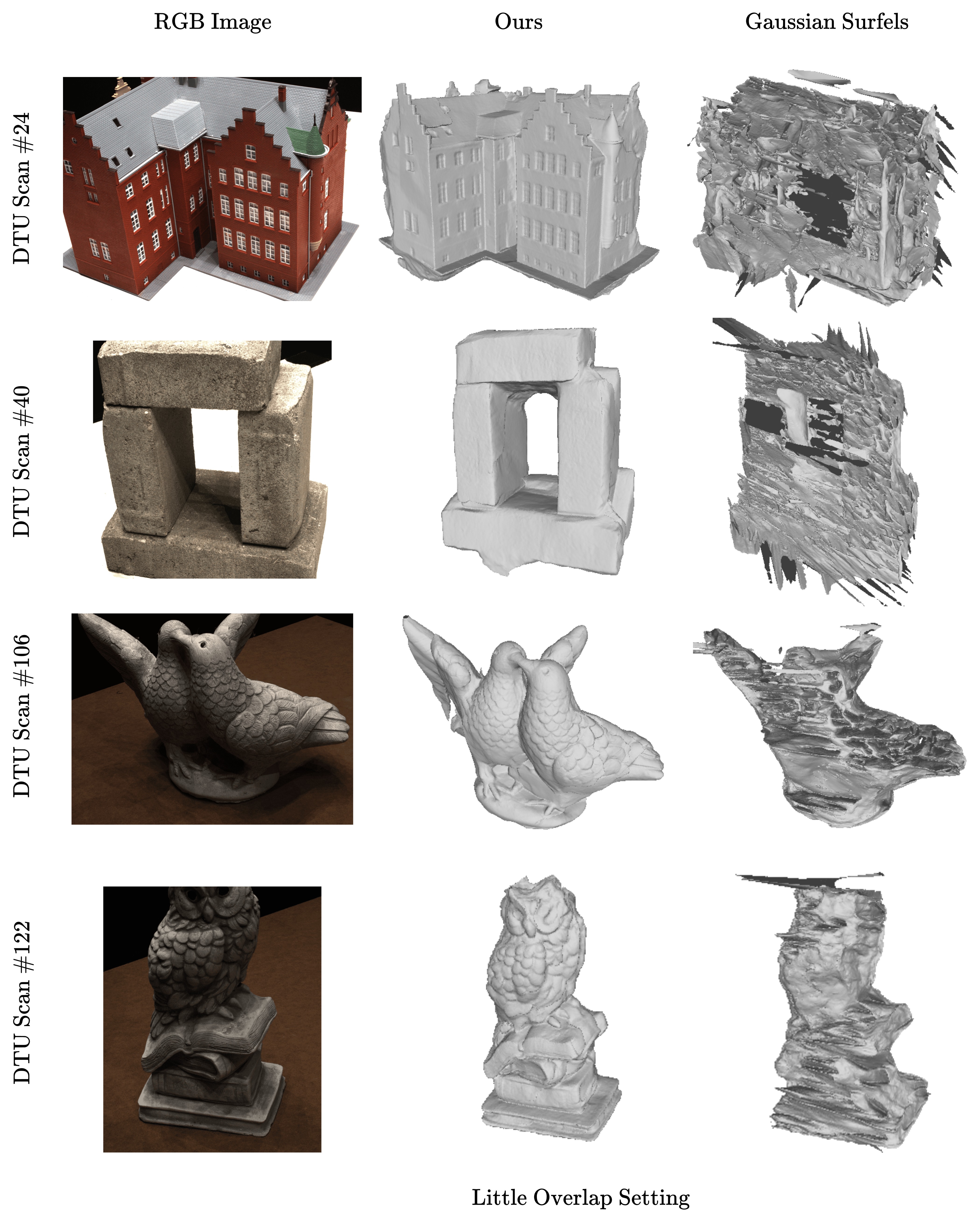}}
    \caption{Qualitative comparison between our method and Gaussian Surfels \cite{Dai2024GaussianSurfels} on the DTU dataset under little overlap setting.}
    \label{fig:DTU_little_comp}
\end{figure}

\begin{figure}
    \centering
    \scalebox{0.9}{\includegraphics[width=0.8\textwidth]{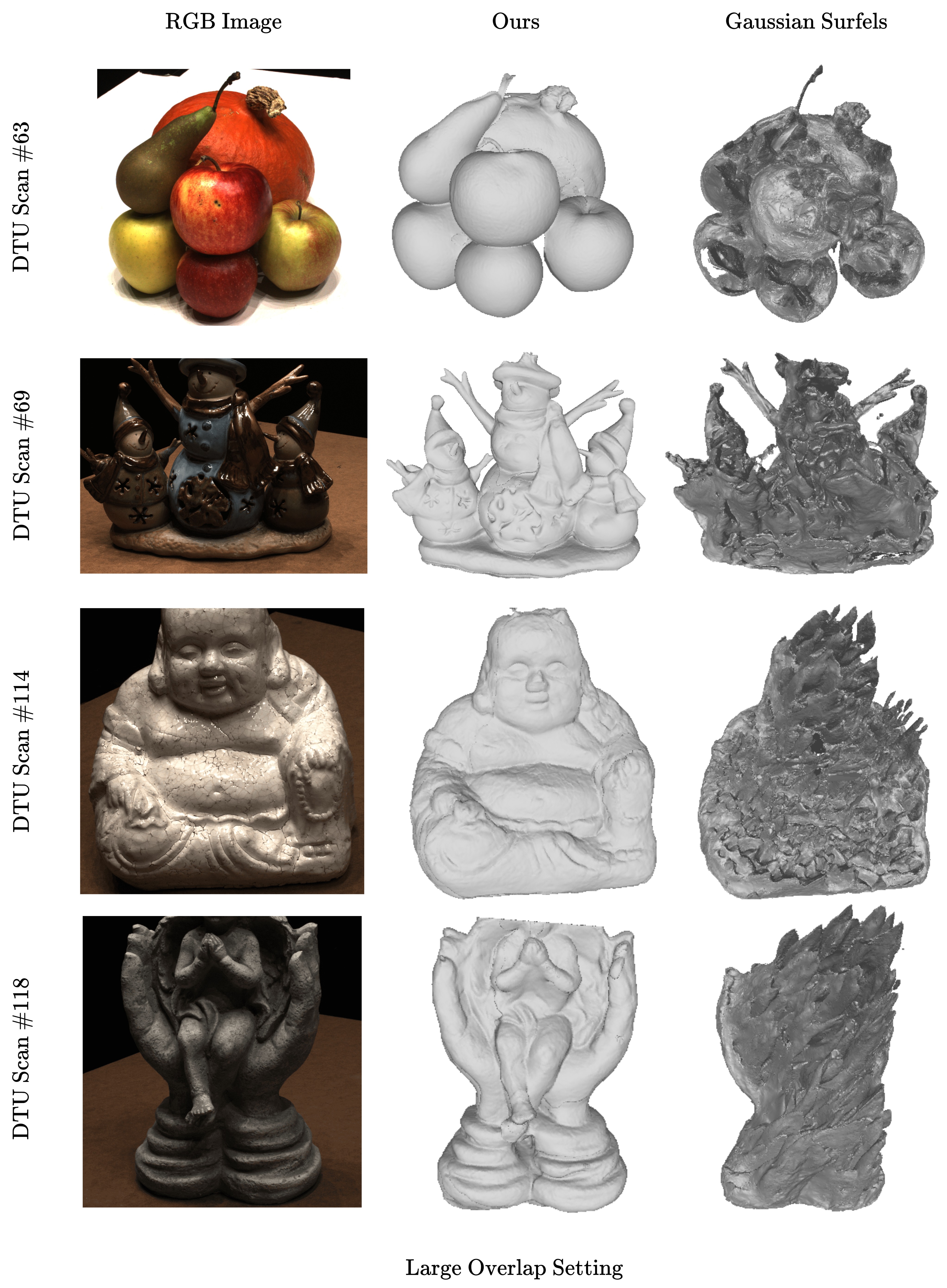}}
    \caption{Qualitative comparison between our method and Gaussian Surfels \cite{Dai2024GaussianSurfels} on the DTU dataset under large overlap setting.}
    \label{fig:DTU_large_comp}
\end{figure}

~\Cref{fig:DTU_little_comp} and~\ref{fig:DTU_large_comp} provide a visual comparison of the results of our method with the results of Gaussian Surfels~\cite{Dai2024GaussianSurfels}. As one can see, Gaussian Surfels significantly fail in sparse multiview stereo, both in the little-overlap and large-overlap settings. Our method, on the other hand, recovers highly accurate 3D geometry in both settings.

\end{document}